\documentclass[preprint,12pt,authoryear]{elsarticle}
\usepackage{amssymb}
\usepackage{amsmath}
\usepackage[table,xcdraw]{xcolor}
\usepackage{graphicx}
\usepackage{mathrsfs}
\usepackage{hyperref}
\usepackage{tikz}
\usetikzlibrary{arrows.meta, positioning}
\usepackage{algorithm}
\usepackage{algpseudocode}
\usepackage{txfonts}
\usepackage{booktabs} 
\usepackage{makecell} 
\usepackage{multirow}
\usepackage{changepage}
\usepackage{rotating}
\usepackage{longtable}
\usepackage{subcaption} 
\usepackage{listings}
\usepackage{bm} 
\usepackage{array}  
\usepackage{etoolbox} 

\usepackage{draftwatermark}
\SetWatermarkText{PREPRINT}
\SetWatermarkScale{0.7}
\SetWatermarkColor[gray]{0.90}
\SetWatermarkAngle{45}

\journal{N/A}

\begin{document}

\begin{frontmatter}

\title{Causal Discovery in Multivariate Time Series through Mutual Information Featurization}
\author[label1]{Gian Marco Paldino}
\ead{gian.marco.paldino@ulb.be}
\author[label1]{Gianluca Bontempi}

\affiliation[label1]{organization={Machine Learning Group, Computer Science Department, Universite' Libre de Bruxelles},
            addressline={Bld de Triomphe}, 
            city={Ixelles},
            postcode={1050}, 
            state={Brussels},
            country={Belgium}}

\begin{abstract}
Discovering causal relationships in complex multivariate time series is a fundamental scientific challenge. Traditional methods often falter, either by relying on restrictive linear assumptions or on conditional independence tests that become uninformative in the presence of intricate, non-linear dynamics. This paper proposes a new paradigm, shifting from statistical testing to pattern recognition. We hypothesize that a causal link creates a persistent and learnable asymmetry in the flow of information through a system's temporal graph, even when clear conditional independencies are obscured. We introduce Temporal Dependency to Causality (TD2C), a supervised learning framework that operationalizes this hypothesis. TD2C learns to recognize these complex causal signatures from a rich set of information-theoretic and statistical descriptors. Trained exclusively on a diverse collection of synthetic time series, TD2C demonstrates remarkable zero-shot generalization to unseen dynamics and established, realistic benchmarks. Our results show that TD2C achieves state-of-the-art performance, consistently outperforming established methods, particularly in high-dimensional and non-linear settings. By reframing the discovery problem, our work provides a robust and scalable new tool for uncovering causal structures in complex systems.
\end{abstract}


\end{frontmatter}

\section{Introduction}
\label{sec:introduction}

The pursuit of understanding causal relationships within time series data represents a cornerstone of scientific inquiry across diverse disciplines. Traditionally, this field has been dominated by methods like Granger causality \citep{granger1980testing}, which, despite its widespread use, relies on linear assumptions and temporal precedence, often mistaking correlation for causation. While more advanced bivariate methods like the Additive Noise Model (ANM; \citealp{hoyer2008nonlinear}) and Information Geometric Causal Inference (IGCI; \citealp{daniusis2012inferring}) have been developed, their applicability is limited in real-world systems where complex, multivariate interactions are the norm.

To address this, the focus has shifted towards multivariate approaches capable of handling confounding. Constraint-based methods like PCMCI \citep{runge2019detecting}, noise-based methods such as VarLiNGaM \citep{hyvarinen2010estimation}, and score-based techniques like DYNOTEARS \citep{pamfil2020dynotears} have emerged as powerful tools. 

This paper introduces a novel perspective on this challenge.  We introduce a novel hypothesis: a causal link creates a fundamental and learnable asymmetry in the flow of information through the system's temporal graph. In simple cases, this asymmetry manifests as a clear conditional independence that can be detected by independence tests. However, in the complex, confounded scenarios typical of real-world data, this binary signal is lost. We posit that a more subtle, quantitative signature of causality persists. This signature is not a simple zero/non-zero value but is encoded in the statistical distribution of information-theoretic quantities, reflecting an underlying asymmetry in the number and nature of open information pathways.

We operationalize this hypothesis with our proposed method, TD2C (Temporal Dependency to Causality). Building on the supervised learning paradigm of the static Dependency to Causality (D2C) framework \citep{bontempi2015dependency}. TD2C learns to recognize the complex, asymmetric patterns in a rich set of features that signify a causal connection. By engineering descriptors that explicitly capture temporal dynamics and information flow, TD2C is trained on synthetic data to build a model of causality that can be deployed to make predictions on new, unseen systems.
 
The main contributions of this work are the following:
\begin{itemize}
    \item We propose a new theoretical grounding for data-driven causal discovery in time series, based on the hypothesis that causal directionality creates a persistent, learnable asymmetry in the distribution of information-theoretic quantities, even when simple independence tests fail.
    \item We introduce TD2C, a supervised learning framework that operationalizes this hypothesis by featurizing the problem in a way that captures the unique signatures of temporal dynamics.
    \item We conduct a comprehensive benchmark against state-of-the-art methods, demonstrating that TD2C is a competitive and robust approach, particularly in complex and high-dimensional settings.
\end{itemize}
 
The remainder of the manuscript is organized as follows. Section \ref{sec:sota} reviews related work in causal discovery. Section \ref{sec:background} provides the necessary background on information theory and the foundational D2C approach. Section \ref{sec:contribution} details our proposed method, TD2C, and the theoretical motivation behind it. Section \ref{sec:experiments} presents the experimental setup and Section \ref{sec:results} discusses the corresponding results. Section \ref{sec:conclusion_and_future_work} concludes the work, discussing limitations and providing future research directions.

\section{State of the art}\label{sec:sota}

The literature on causal discovery from time series data is vast and encompasses a wide array of methodologies, each designed to tackle various aspects of the problem under different assumptions. These methodologies can be broadly classified into several families \citep{assaad2022survey}, each with distinct characteristics and applications. For a more comprehensive review of these methods and to understand the full scope of the field, we invite readers to consult the detailed discussions available in the literature \citep{assaad2022survey}.

\subsection{Granger-based Methods}
This family of methods is based on the idea that past values of one variable help predict the current value of another variable if they are causally connected. The simplest implementation is the Pairwise Granger causality test \citep{granger1980testing}, whose null hypothesis posits that $\mathbf{z}_i$ does not Granger cause $\mathbf{z}_j$. A statistical test is performed to demonstrate that the inclusion of past values of $\mathbf{z}_i$ significantly enhances the prediction of $\mathbf{z}_j$ when past values of $\mathbf{z}_j$ are also used as regressors. If the resulting p-values from the test are below the threshold set for statistical significance, the null hypothesis is rejected. 

\subsection{Constraint-Based Methods}
These methods rely on statistical tests for conditional independence to infer causal relationships from the data, and build a causal graph by systematically adding or removing edges based on the independence tests. The Peter and Clark algorithm \citep{spirtes2001causation} performs conditional independence tests between pair of variables, using increasingly larger conditioning sets. In a large-variate time series context, when lagged variables need also to be tested, the problem becomes easily intractable, especially for all possible combinations $q$ of conditioning sets. To tackle this problem, more advanced algorithms such as PCMCI (Peter and Clark Momentary Conditional Independence, \citealp{runge2019detecting}) have been proposed. This algorithm is divided into two phases: an initial $PC_{1}$ phase, where a variation of the PCstable \citep{colombo2014order} algorithm is applied, only testing the $p$ parents with strongest dependency, that is, restricting the maximum number of combinations $q_{\max }$ per iteration to $q_{\max }=1$. The output of $PC_{1}$ is a superset of parents $\widehat{\mathcal{P}}\left(z_j^t\right)$ for each variable $\mathbf{z}_j$ at each time-step $t$, thus removing irrelevant conditions for each of the $N$ variables by iterative independence testing. This step is followed by the MCI phase. Here, for each candidate link $z_i^{t-\tau} \rightarrow z_j^t$, a Momentary Conditional Independence (MCI) test is performed to check for the conditional dependence
    \begin{equation}\label{pcmci}
    z^{t-\tau}_i \not \Perp z^t_j \mid \widehat{\mathcal{P}}\left(z^t_j\right) \backslash\left\{z^{t-\tau}_i\right\}, \widehat{\mathcal{P}}\left(z^{t-\tau}_i\right),
    \end{equation}
conditioning on both the parents of $z^t_j$ and the time-shifted parents of $z^{t-\tau}_i$. This phase addresses false positive control for the highly interdependent time series case. Variations have been proposed to address contemporaneous links \citep{runge2020discovering} and the presence of latent confounders \citep{gerhardus2020high}.

\subsection{Noise-Based Approaches}
They exploit the non-Gaussian nature of the data to distinguish between cause and effect based on the independence of the residuals (noise) when one variable is regressed on another.  
Considering a system where each variable in $ Z$ is linearly influenced by all other variables in $Z$ through the coefficients matrix $ B $, plus an independent non-Gaussian noise component $ e $, we can write $Z=BZ+e$. If we define $A=(I-B)^{-1}$, this can be written as $Z=Ae$. If the available data Z is a linear, invertible mixture of non-Gaussian independent disturbance variables, it can be shown that the mixing matrix A is identifiable and this process defines the standard linear Independent Component Analysis (ICA, \citealp{comon1994independent}) model. Solving for $A$ through ICA is the core aspect of the LiNGAM \citep{shimizu2006linear} family of methods, whose extensions to time series data include VARLiNGAM \citep{hyvarinen2010estimation}, where the temporal dynamics are modeled as 
\begin{equation}
Z^t=\sum_{\tau=1}^l B_\tau Z^{t-\tau}+e_t
\end{equation}
where $l$ is the number of time-delays used, that is, the order of the autoregressive model, $B_\tau$, $\tau= 1$, $\ldots$, $l$ are $n \times n$ matrices, and $e_t$ is the non-gaussian disturbance.

\subsection{Score-Based Strategies}
These techniques use a scoring criterion to evaluate and compare different causal models. For example, DYNOTEARS \citep{pamfil2020dynotears} frames the problem of causal discovery as the following optimization problem:
\begin{equation}
\min _{W, A} \frac{1}{2 n}\|Z^t - Z^t W - Y A\|_F^2 . \text { s.t. } W \text { is acyclic}.
\end{equation}
Here, $\left\|\cdot\right\|_F^2$ denotes the squared Frobenius norm, $Z^t$ is the input time series vector, $Y$ is the lagged time series vector, and $W$ and $A$ are the contemporaneous and time-lagged weighted adjacency matrices, respectively, to be estimated. 

\subsection{Recent Benchmark-Optimized Methods}
It is essential to also highlight recent algorithmic developments that have demonstrated strong performances in public benchmarks, such as SLARAC (Subsampled Linear Auto-Regression Absolute Coefficients), QRBS (Quantiles of Ridge regressed Bootstrap Samples), LASAR (Lasso Auto-Regression), and SELVAR (Selective auto-regressive model) \citep{weichwald2020causal}. However, as pointed out by \citet{reisach2021beware}, data scale and marginal variance may carry information about the data generating process. These methods might leverage this to dominate benchmarking results, such as, for example, the outcome of the NeurIPS Causality 4 Climate competition.

\section{Background: the D2C approach}\label{sec:background}
This section provides the basics for understanding the Dependency-to-Causality (D2C) framework from \citep{bontempi2015dependency} for causal discovery from observational data. As in \citep{bontempi2015dependency}, we denote random variables by boldface letters (e.g., $\mathbf{Z}$), while their realizations are in lowercase (e.g., $z$). Subscripts are used to identify specific variables within a set (e.g., $\mathbf{z}_i$), and superscripts denote time steps (e.g., $z_i^{(t)}$). Throughout this section, we consider continuous random variables from an $n$-variate distribution $\mathbf{Z} = \left[\mathbf{z}_1, \ldots, \mathbf{z}_n \right]$ with a joint Lebesgue density.

The D2C approach leverages information-theoretic measures, particularly mutual information, to construct asymmetric descriptors that can distinguish causal direction between pairs of variables. The fundamental insight is that causal relationships exhibit asymmetric patterns in their information-theoretic properties when conditioning on different subsets of variables from their Markov Blankets. This section establishes the theoretical foundation of this approach, beginning with the necessary background on mutual information (Section \ref{sec:background-mi-intro}) and causal graphical models (Section \ref{sec:background-cgm}), before detailing how asymmetric descriptors are constructed and utilized for causal inference (Section \ref{sec:background-descriptors}). We then discuss the mutual information estimation approach used in the original D2C framework (Section \ref{sec:background-mi}) and the assumptions required for the static case (Section \ref{sec:background-assumptions}), concluding with the preliminary extension to temporal causal discovery (Section \ref{sec:background-cad2c}).

\subsection{Mutual Information}\label{sec:background-mi-intro}

The mutual information $I(\mathbf{z}_i; \mathbf{z}_j)$ quantifies the statistical dependency between two random variables $\mathbf{z}_i$ and $\mathbf{z}_j$. It is defined as the expected value of the logarithmic ratio between the joint and marginal probability density functions, as shown in \eqref{eq:mutual-information-two-variables}. The conditional mutual information, $I(\mathbf{z}_i ; \mathbf{z}_j \mid \mathbf{z}_k)$, extends this concept to a conditional setting \eqref{eq:mutual-information-three-variables}.

\begin{equation}\label{eq:mutual-information-two-variables}
I\left(\mathbf{z}_i ; \mathbf{z}_j\right) = \mathbb{E}_{\mathbf{z}_i, \mathbf{z}_j}\left[\log \frac{p\left(z_i, z_j\right)}{p\left(z_i\right) p\left(z_j\right)}\right]
\end{equation}

\begin{equation}\label{eq:mutual-information-three-variables}
I\left(\mathbf{z}_i ; \mathbf{z}_j \mid \mathbf{z}_k\right) = \mathbb{E}_{\mathbf{z}_i, \mathbf{z}_j, \mathbf{z}_k}\left[\log \frac{p\left(z_i, z_j \mid z_k\right)}{p\left(z_i \mid z_k\right) p\left(z_j \mid z_k\right)}\right]
\end{equation}

Here, $p(\cdot)$ denotes a probability density function, and $\mathbb{E}_{\mathbf{z}_i, \dots}$ denotes the expectation taken over the joint distribution of the subscripted random variables. These definitions are equivalent to the integral form, assuming the convention $0 \log \frac{0}{0}=0$.

Mutual information quantifies the amount of information obtained about one random variable by observing another. Formally, the mutual information between $\mathbf{z}_i$ and $\mathbf{z}_j$ is zero if and only if these variables are statistically independent, i.e., $I(\mathbf{z}_i ; \mathbf{z}_j) = 0 \iff \mathbf{z}_i \Perp \mathbf{z}_j$. This relationship underscores the utility of mutual information as a measure of dependence.

\subsection{Causal Graphical Models and Information Theory}\label{sec:background-cgm}

A directed graph $\mathcal{G}$ is a pair $(\mathcal{V}, \mathcal{E})$ where $\mathcal{V}$ is a finite non-empty set whose elements are called nodes, and $\mathcal{E}$ is a set of ordered pairs of distinct elements of $\mathcal{V}$. The elements of $\mathcal{E}$ are called edges. A directed graph is called a directed acyclic graph (DAG) if it contains no directed cycles, i.e. if its nodes don't form a directional closed loop. DAGs are an effective way of representing multivariate distributions where the nodes denote random variables, and the topology encodes conditional independence assertions, i.e. two non-connected nodes are conditionally independent \citep{pearl2009causality}.

The Markov Blanket $\mathbf{M}_i$ of a variable $\mathbf{z}_i$ is the set of variables that shields the rest of the system from that specific variable \citep{koller2009probabilistic}. In other words, when conditioning on $\mathbf{M}_i$, $\mathbf{z}_i$ becomes conditionally independent of all the remaining variables. This means that it is possible to describe this structural notion in terms of conditional mutual information as follows: $\mathbf{M}_i$ is the smallest subset of variables from $\mathbf{Z} \setminus \{\mathbf{z}_i\}$ such that:

\begin{equation}
I\left(\mathbf{z}_i ;\left(\mathbf{Z} \backslash\left(\mathbf{M}_i \cup \{\mathbf{z}_i\}\right)\right) \mid \mathbf{M}_i\right)=0
\end{equation}

This formulation is equivalent to:
$
   \mathbf{M}_i = \mathbf{C}_i \cup \mathbf{E}_i \cup \mathbf{S}_i
$
where $\mathbf{C}_i = \{ \mathbf{z}_c | \mathbf{z}_c \rightarrow \mathbf{z}_i \} $ is the set of variables that have direct arrows pointing to $\mathbf{z}_i$ (causes), $\mathbf{E}_i = \{ \mathbf{z}_e | \mathbf{z}_i \rightarrow \mathbf{z}_e \} $, are the set of variables to which $\mathbf{z}_i$ has a direct causal influence, i.e., where $\mathbf{z}_i$ is the parent (effects), and $\mathbf{S}_i = \{ \mathbf{z}_s | \exists \mathbf{z}_j : \mathbf{z}_s \rightarrow \mathbf{z}_j \leftarrow \mathbf{z}_i \text{ and } \mathbf{z}_s \neq \mathbf{z}_i \}$, are the variables that are not direct causes or effects of $\mathbf{z}_i$ but are connected to $\mathbf{z}_i$ through a common effect (spouses). We denote members of $\mathbf{C}_i$, $\mathbf{E}_i$, $\mathbf{S}_i$ as $\mathbf{c}_{i,k_c}$, $\mathbf{e}_{i,k_e}$, $\mathbf{s}_{i,k_s}$, respectively, where $k_c$, $k_e$, $k_s$ are indices for multiple elements from each set.

\begin{figure}[!ht]
\centering
\begin{tikzpicture}[
    node distance=1.5cm,
    greenroundnode/.style={circle, draw=black, fill=green!20, minimum size=8mm},
    grayroundnode/.style={circle, draw=black, fill=gray!20, minimum size=8mm},
    roundnode/.style={circle, draw=black, fill=none, minimum size=8mm},
    ]

        \node[greenroundnode]    (zi)                              {$\mathbf{z}_i$};
    \node[greenroundnode]    (zj)       [below right=of zi]    {$\mathbf{z}_j$};
    \node[roundnode]         (c1i)      [above left=of zi]     {$\mathbf{c}_{i,1}$};
    \node[grayroundnode]     (l)        [left=of c1i]          {$\mathbf{l}$};
    
    \node[roundnode]         (c2i)      [above =of zi]         {$\mathbf{c}_{i,2}$};
    \node[roundnode]         (e1i)      [below=of zi]          {$\mathbf{e}_{i,1}$};
    \node[roundnode]         (s1i)      [above left=of e1i]    {$\mathbf{s}_{i,1}$};
    \node[grayroundnode]     (m)        [left=of s1i]          {$\mathbf{m}$};
    
    \node[roundnode]         (c1j)      [above=of zj]          {$\mathbf{c}_{j,1}$};  
    \node[roundnode]         (c2j)      [above right=of zj]    {$\mathbf{c}_{j,2}$};
    \node[grayroundnode]     (n)        [right=of c2j]         {$\mathbf{n}$};
    
    \node[roundnode]         (e1j)      [below=of zj]          {$\mathbf{e}_{j,1}$};
    \node[roundnode]         (s1j)      [above right=of e1j]   {$\mathbf{s}_{j,1}$};
    \node[grayroundnode]     (o)        [right=of s1j]         {$\mathbf{o}$};
    
        \draw[->, very thick] (zi) -- (zj);
    
    \draw[->] (c1i) -- (zi);
    \draw[->] (c2i) -- (zi);
    \draw[->] (c1j) -- (zj);
    \draw[->] (c2j) -- (zj);
    
    \draw[->] (zi) -- (e1i);
    \draw[->] (zj) -- (e1j);
    \draw[->] (s1j) -- (e1j);
    \draw[->] (s1i) -- (e1i);

    \draw[->] (l) -- (c1i);
    \draw[->] (m) -- (s1i);
    \draw[->] (n) -- (c2j);
    \draw[->] (o) -- (s1j);

\end{tikzpicture}
\caption{The figure illustrates the concept of a Markov Blanket within a Directed Acyclic Graph (DAG). The DAG showcases a section of a larger network, highlighting the variables $\mathbf{z}_i$ and $\mathbf{z}_j$ and their causal connection (thicker arrow). The members of the Markov Blanket for each variable are shown with a white background. These include the parents ($\mathbf{c}_{i,1}, \mathbf{c}_{i,2}$ and $\mathbf{c}_{j,1},\mathbf{c}_{j,2}$), children ($\mathbf{e}_{i,1}$ and $\mathbf{e}_{j,1}$), and spouses ($\mathbf{s}_{i,1}$ and $\mathbf{s}_{j,1}$). The variables $\mathbf{l}, \mathbf{m}, \mathbf{n}, \mathbf{o}$ (gray background) represent other nodes in the network.}
\label{fig:dag}
\end{figure}
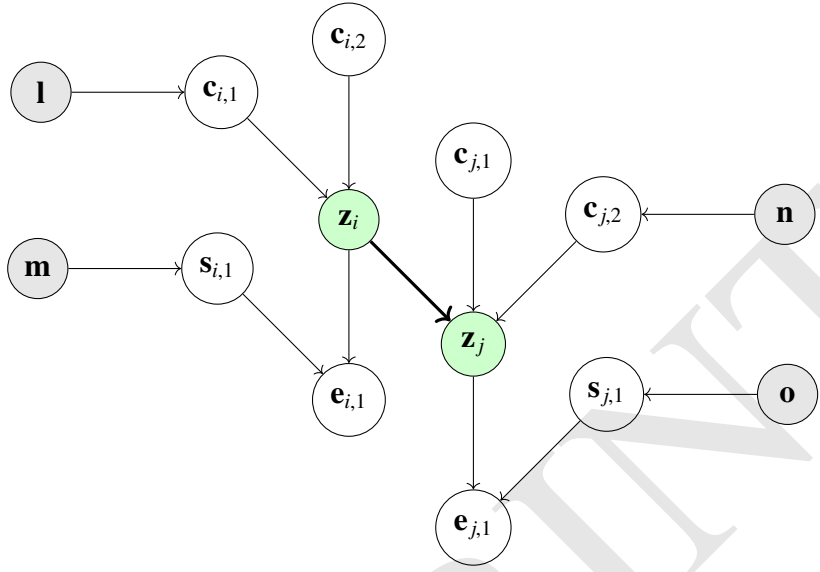

Figure \ref{fig:dag} illustrates the concept of a Markov Blanket and its components within a Directed Acyclic Graph (DAG). The DAG depicted showcases a section of a more extensive network, highlighting the variables $\mathbf{z}_i$ and $\mathbf{z}_j$ and their causal connection (thicker arrow). The Markov Blankets for both $\mathbf{z}_i$ and $\mathbf{z}_j$ are reported: parents $\mathbf{c}_{i,1}, \mathbf{c}_{i,2}$ and $\mathbf{c}_{j,1},\mathbf{c}_{j,2}$, children $\mathbf{e}_{i,1}$ and $\mathbf{e}_{j,1}$, and spouses $\mathbf{s}_{i,1}$ and $\mathbf{s}_{j,1}$, are specified. The number of variables in the Markov blanket can vary. Variables $\mathbf{l}$, $\mathbf{m}$, $\mathbf{n}$, $\mathbf{o}$ are additional variables that represent the remaining network.

Alternatively, these causal connections can also be represented using Structural Equation Models (SEMs) \citep{bielby1977structural}. \eqref{eq:SEM-static} represents the DAG in Figure \ref{fig:dag} assuming a linear Additive Noise Model \citep{hoyer2008nonlinear}.

\begin{equation}\label{eq:SEM-static}
    \begin{aligned}
    \mathbf{z}_i &= \alpha_1 \mathbf{c}_{i,1} + \alpha_2 \mathbf{c}_{i,2} + \alpha_3 \mathbf{s}_{i,1} + \boldsymbol{\epsilon}_i \\
    \mathbf{z}_j &= \beta_1 \mathbf{z}_i + \beta_2 \mathbf{c}_{j,1} + \beta_3 \mathbf{c}_{j,2} + \beta_4 \mathbf{s}_{j,1} + \boldsymbol{\epsilon}_j \\
    \mathbf{c}_{i,1} &= \gamma_1 \mathbf{l} + \boldsymbol{\epsilon}_{c_{i,1}} \\
    \mathbf{s}_{i,1} &= \delta_1 \mathbf{m} + \boldsymbol{\epsilon}_{s_{i,1}} \\
    \mathbf{c}_{j,2} &= \eta_1 \mathbf{n} + \boldsymbol{\epsilon}_{c_{j,2}} \\
    \mathbf{s}_{j,1} &= \theta_1 \mathbf{o} + \boldsymbol{\epsilon}_{s_{j,1}}
    \end{aligned}
\end{equation}

where \( \alpha_1, \alpha_2, \alpha_3, \beta_1, \beta_2, \beta_3, \beta_4, \gamma_1, \delta_1, \eta_1, \theta_1 \) are coefficients representing the strengths of the causal relationships between variables and \( \boldsymbol{\epsilon}_i, \boldsymbol{\epsilon}_j, \dots \) represent the error terms or disturbances in the model, which are also random variables capturing the influences of unobserved factors.

SEMs describe the relationships among variables using a set of equations, where each equation represents how a variable is generated from its direct causes plus some error term. These causal representations facilitate the understanding of how information flows through the system and how interventions might impact the system's behavior. As illustrated in Figure \ref{fig:dag}:
\begin{itemize}
    \item $ \mathbf{c}_{i,1} \rightarrow \mathbf{z}_i \rightarrow \mathbf{z}_j $ is called a \textit{chain}: when conditioning on $\mathbf{z}_i$, $\mathbf{c}_{i,1}$ and $\mathbf{z}_j$ become independent, because the mediator $\mathbf{z}_i$ is observed.
    \item The structure $ \mathbf{e}_{i,1} \leftarrow \mathbf{z}_i \rightarrow \mathbf{z}_j $ is called a \textit{fork}: when we condition on $\mathbf{z}_i$, $\mathbf{e}_{i,1}$ and $\mathbf{z}_j$ become independent due to the common cause $\mathbf{z}_i$ being observed.
    \item The structure $ \mathbf{z}_i \rightarrow \mathbf{z}_j \leftarrow \mathbf{c}_{j,1} $ is called a \textit{collider}: conditioning on $\mathbf{z}_j$ (or on a descendant of $\mathbf{z}_j$) introduces a dependence between $\mathbf{z}_i$ and $\mathbf{c}_{j,1}$ because it opens a path through the common effect $\mathbf{z}_j$.
\end{itemize}

\subsection{Inferring Directionality with Asymmetric Descriptors}\label{sec:background-descriptors}

The ultimate goal is to infer a DAG from observational data, which requires determining not only the presence of dependencies between variables but also their causal direction. A dependency descriptor of the pair $(\mathbf{z}_i, \mathbf{z}_j)$ is a function $d(i,j)$ of the distribution of $\mathbf{Z}$, symmetric if $d(i,j) = d(j,i)$, asymmetric otherwise \citep{wiedermann2020direction}. Examples of symmetric descriptors include the correlation $\rho(\mathbf{z}_i, \mathbf{z}_j)$ or the mutual information $I(\mathbf{z}_i; \mathbf{z}_j)$. Conditional mutual information, on the other hand, allows for the creation of several asymmetric descriptors. In the context of Figure \ref{fig:dag}, consider the structural configurations given by $\mathbf{z}_i \rightarrow \mathbf{z}_j \leftarrow \mathbf{c}_{j,1}$ (a collider) and $\mathbf{c}_{i,2} \rightarrow \mathbf{z}_i \rightarrow \mathbf{z}_j$ (a chain). When conditioning on $\mathbf{z}_j$, a dependency between $\mathbf{z}_i$ and its spouse $\mathbf{c}_{j,1}$ emerges, whereas conditioning on $\mathbf{z}_i$ makes its cause $\mathbf{c}_{i,2}$ independent of its effect $\mathbf{z}_j$. This can be written as:

\begin{equation}
\mathbf{z}_i \rightarrow \mathbf{z}_j \iff \mathbf{z}_i \not\Perp \mathbf{c}_{j,k} \mid \mathbf{z}_j \text{ and } \mathbf{z}_j \Perp \mathbf{c}_{i,k} \mid \mathbf{z}_i \quad \forall k
\end{equation}

From this approach, it is possible to derive a collection of (conditional) mutual information terms, such as $I\left(\mathbf{z}_i ; \mathbf{c}_{j,k} \mid \mathbf{z}_j\right)$, or $I\left(\mathbf{e}_{i,k} ; \mathbf{c}_{j,l} \mid \mathbf{z}_j\right)$. These quantities are greater than zero, while their counterparts with $i$ and $j$ swapped are null. Idealized asymmetric descriptors can be computed with similar reasoning:
$$
\begin{gathered}
d_{a,k}(i, j)=I\left(\mathbf{z}_i ; \mathbf{c}_{j,k} \mid \mathbf{z}_j\right) > 0, \quad d_{a,k}(j, i)=I\left(\mathbf{z}_j ; \mathbf{c}_{i,k} \mid \mathbf{z}_i\right) = 0 \\
d_{b,k,l}(i, j)=I\left(\mathbf{e}_{i,k} ; \mathbf{c}_{j,l} \mid \mathbf{z}_j\right) > 0, \quad d_{b,k,l}(j, i)=I\left(\mathbf{e}_{j,l} ; \mathbf{c}_{i,k} \mid \mathbf{z}_i\right) = 0 \\
d_{c,k,l}(i, j)=I\left(\mathbf{c}_{i,k} ; \mathbf{c}_{j,l} \mid \mathbf{z}_j\right) > 0, \quad d_{c,k,l}(j, i)=I\left(\mathbf{c}_{j,l} ; \mathbf{c}_{i,k} \mid \mathbf{z}_i\right) = 0 \\
d_{d,k}(i, j)=I\left(\mathbf{z}_j ; \mathbf{c}_{i,k}\right) > 0, \quad d_{d,k}(j, i)=I\left(\mathbf{z}_i ; \mathbf{c}_{j,k}\right) = 0
\end{gathered}
$$

As noted in \citep{bontempi2015dependency}, the idealized descriptors above require pre-identifying the members of a Markov Blanket as causes, effects, or spouses—a 'vicious circle' that requires the very causal knowledge we seek. The innovation of D2C is to bypass this by analyzing the statistical properties of descriptor populations. Rather than relying on pre-classified variables, the framework computes descriptor terms for \textit{every} member $\mathbf{m}_{j,k}$ of the relevant Markov Blanket. This generates a population of values which is an unknown mixture of terms related to causes, effects, and spouses. The key insight is that the properties of the underlying subpopulations create a detectable asymmetry in the \textit{overall mixture}. For the true causal link $\mathbf{z}_i \rightarrow \mathbf{z}_j$, the population of descriptors associated with the pair $(i,j)$ will be statistically different from the population associated with the reverse pair $(j,i)$. For instance, the distribution of $\{I\left(\mathbf{z}_i ; \mathbf{m}_{j,k} \mid \mathbf{z}_j\right)\}_{k}$ is skewed towards positive values, while the distribution of $\{I\left(\mathbf{z}_j ; \mathbf{m}_{i,k} \mid \mathbf{z}_i\right)\}_{k}$ is concentrated at zero. To characterize this distributional asymmetry, D2C uses a set of quantiles of the empirical distributions of these terms. These quantiles serve as robust features that summarize the entire mixture's characteristics. The rationale is that if $\mathbf{z}_i \rightarrow \mathbf{z}_j$, the vector of quantiles computed for $(i,j)$ will be systematically different from the vector computed for $(j,i)$, providing an effective basis for classification.

The primary descriptors are constructed from three families of conditional and unconditional mutual information terms. For a potential causal link from $\mathbf{z}_i$ to $\mathbf{z}_j$, these populations are:

\begin{align}
\mathcal{D}_1(i,j) &= \left\{ I(\mathbf{z}_i; \mathbf{m}_{j,k} \mid \mathbf{z}_j) \right\}_{k=1, \dots, |\mathbf{M}_j|} \label{eq:d1_desc}\\
\mathcal{D}_2(i,j) &= \left\{ I(\mathbf{m}_{i,k}; \mathbf{m}_{j,l} \mid \mathbf{z}_j) \right\}_{k=1, \dots, |\mathbf{M}_i|, l=1, \dots, |\mathbf{M}_j|} \label{eq:d2_desc}\\
\mathcal{D}_3(i,j) &= \left\{ I(\mathbf{z}_j; \mathbf{m}_{i,k}) \right\}_{k=1, \dots, |\mathbf{M}_i|} \label{eq:d3_desc}
\end{align}

\subsection{Estimating Mutual Information}\label{sec:background-mi}
In \citep{bontempi2015dependency} and \citep{bontempi2020learning}, conditional mutual information is estimated using the identity:

\begin{equation}\label{eq:MI-estimate}
I\left(\mathbf{z}_1 ; \mathbf{z}_2 \mid \mathbf{z}_3\right)=H\left(\mathbf{z}_1 \mid \mathbf{z}_3\right)-H\left(\mathbf{z}_1 \mid \mathbf{z}_2, \mathbf{z}_3\right)
\end{equation}
where entropy terms are inferred from the normalized mean squared error (NMSE) of predictions made by a Ridge regression model. This approach indirectly measures how much information the predictor variables contain about the outcome.

This mutual information estimator relies on the assumption that all conditional distributions are Gaussian, which can lead to severe underestimation when this assumption is violated. Furthermore, Ridge regression fails to detect nonlinear dependencies between variables—a particularly problematic limitation in complex systems where causal relationships are often nonlinear. These limitations suggest that while this approach offers computational efficiency for approximately linear, Gaussian systems, it may miss important causal relationships in real-world time series that exhibit nonlinearity or non-Gaussianity.

\subsection{Assumptions for the Static Case}\label{sec:background-assumptions}
To infer a DAG from observational data, it is essential to make specific assumptions. For example, in the context of Figure \ref{fig:dag}, consider the structural configuration given by $\mathbf{z}_j \leftarrow \mathbf{z}_i \rightarrow \mathbf{e}_{i,1}$. If $\mathbf{z}_i$ was not observed, one would infer a spurious correlation between $\mathbf{z}_j$ and $\mathbf{e}_{i,1}$. A set of variables is said to be causally sufficient if all common causes of all variables in the set are observed \citep{spirtes2001causation}. A necessary and sufficient condition for a probability distribution to be compatible with a DAG is the independence of every variable from its non-descendants when conditioning on its parents. This is referred to as the Markov Condition \citep{pearl2000models}. We say that a DAG and a compatible probability distribution are faithful to one another if the Markov condition applied to the DAG entails all (and only) the true conditional independence relations in the distribution \citep{spirtes2001causation, assaad2022survey}. In the following, we assume causal sufficiency, the Markov condition, and faithfulness.

\subsection{Transition to Temporal Causal Discovery}\label{sec:background-cad2c}
A preliminary attempt to extend D2C to the temporal case was made by \citep{bontempi2020learning}. However, this initial extension was limited, essentially treating the temporal problem as a static case by considering lagged variables as standalone variables. While this represented a first step, it did not fully exploit the temporal structure of the data. This work also introduced a new descriptor based on Information Interaction \citep{mcgill1954multivariate}, designed to address the "vicious circle" of the original D2C approach. The Information Interaction (II) between two members of a Markov Blanket, $\mathbf{m}_{i,k}$ and $\mathbf{m}_{i,l}$, and the central variable $\mathbf{z}_i$ is defined as:

\begin{align}
\text{II}(\mathbf{m}_{i,k}; \mathbf{m}_{i,l}; \mathbf{z}_i) = I(\mathbf{m}_{i,k}; \mathbf{m}_{i,l}) - I(\mathbf{m}_{i,k}; \mathbf{m}_{i,l} \mid \mathbf{z}_i) \label{eq:info_interaction}
\end{align}

The comparison in that work was conducted against several algorithms, including the Semi-Interleaved HITON-PC \citep{aliferis2003hiton}, IAMB \citep{zhang2010improved}, Fast-IAMB \citep{yang2019three}, GS \citep{edera2014grow}, PC \citep{kalisch2007estimating}, and a Granger test \citep{zeileis2002diagnostic}. These are for the most part static approaches that do not fully account for the temporal dynamics inherent in time series data, representing a strong limitation when assessing performance against the modern literature on temporal causal discovery and motivating a more principled extension.

\section{Proposed Method: The TD2C Approach}\label{sec:contribution}

In this work, we introduce Temporal Dependency to Causality (TD2C), a framework that extends the foundational ideas of D2C \citep{bontempi2015dependency} and caD2C \citep{bontempi2020learning} to the temporal domain. While retaining the core supervised learning paradigm, TD2C introduces three fundamental innovations: a novel, temporally-aware Markov Blanket definition; a robust, non-parametric mutual information estimator; and a powerful, nuanced set of causal descriptors. This creates a comprehensive feature space that captures causal asymmetries across multiple mathematical frameworks—information-theoretic, linear algebraic, statistical moments, and probabilistic dependence modeling—maximizing the classifier's ability to distinguish causal from non-causal relationships. A full list and formal definition of all descriptors used in TD2C can be found in \ref{app:full_descriptors}, Tables \ref{tab:descriptor_formulas_math_p1} and ~\ref{tab:descriptor_formulas_math_p2}.

\subsection{Assumptions for the Dynamic Case}\label{sec:contribution-assumption}
In addition to the standard assumptions of causal sufficiency, the Markov condition, and faithfulness (Section \ref{sec:background-assumptions}), our temporal approach relies on two further assumptions: causal stationarity \citep{runge2018conditional}, where causal mechanisms are constant over time, and that all time series are first-order Markov self-causal ($\mathbf{z}_{i}^{(t-1)} \rightarrow \mathbf{z}_{i}^{(t)}$) \citep{schreiber2000measuring}. Causal stationarity allows the infinite temporal DAG to be modeled with a finite window, making the problem computationally tractable. This is reasonable for many real-world systems operating under stable laws over the observation period. The first-order Markov self-causality assumption allows us to skip the computationally expensive Markov Blanket estimation phase, as we have certainty that $\mathbf{z}_{i}^{(t-1)}$ is a parent of $\mathbf{z}_{i}^{(t)}$, as detailed in Section \ref{sec:contribution-mb}. This is reasonable for most dynamical systems exhibiting temporal continuity. We discuss the limitations of these assumptions in Section \ref{sec:limitation}.

\subsection{A Temporally-Aware Markov Blanket}\label{sec:contribution-mb}
In a time-series context, we leverage the temporal structure to define a stable and causally intuitive Markov Blanket (MB). We build upon the assumption of first-order Markov self-causality ($\mathbf{z}_{i}^{(t-1)} \rightarrow \mathbf{z}_{i}^{(t)}$) \citep{schreiber2000measuring}. From this, we define the Markov Blanket for any variable $\mathbf{z}_i^{(t)}$ as its immediate temporal neighbors:
$$
            \mathbf{M}_{\mathbf{z}_i^{(t)}} = \{\mathbf{z}_i^{(t-1)}, \mathbf{z}_i^{(t+1)}\}
$$
This seemingly simple definition has three profound benefits: it is computationally trivial, bypassing the MB estimation phase; it provides a stable conditioning set, immune to statistical estimation errors; and it provides a canonically ordered set of causally distinct neighbors: a parent and a child. This allows us to move beyond statistical summaries (quantiles) and engineer highly specific, interpretable features.

We acknowledge that this approach represents a deliberate trade-off. By restricting the MB to immediate temporal neighbors, we potentially exclude other causally relevant variables. However, we posit that the certainty of membership for the parent and child—guaranteed by the self-causality assumption—provides a more reliable foundation for inference than the uncertainty inherent in statistical MB estimation.

\subsection{A Non-Parametric Mutual Information Estimator}\label{sec:contribution-mi}
The accuracy of any D2C-style method is fundamentally limited by its mutual information (MI) estimates. The original papers employed a proxy based on Ridge regression, assuming linearity and Gaussianity (Section \ref{sec:background-mi}). To capture non-linear dependencies, we replace this proxy with the non-parametric, k-nearest neighbor (kNN) based Kraskov-Stögbauer-Grassberger (KSG) estimator \citep{kraskov2004estimating}. For two random variables $\mathbf{z}_i$ and $\mathbf{z}_j$, the MI is estimated as:
\begin{equation}
\hat{I}(\mathbf{z}_i; \mathbf{z}_j) = \psi(k) - \frac{1}{N} \sum_{s=1}^{N} [\psi(n_{z_i,s} + 1) + \psi(n_{z_j,s} + 1)] + \psi(N)
\end{equation}
where $\psi(\cdot)$ is the digamma function, $k$ is the number of neighbors, $N$ is the sample size, and $n_{z_i,s}$ and $n_{z_j,s}$ are the number of points within the k-th neighbor distance of the $s$-th sample in the marginal spaces of $\mathbf{z}_i$ and $\mathbf{z}_j$, respectively. For conditional MI, which is central to our causal descriptors, we use the identity:
\begin{equation}
I(\mathbf{z}_i; \mathbf{z}_j | \mathbf{z}_k) = I((\mathbf{z}_i, \mathbf{z}_k); \mathbf{z}_j) - I(\mathbf{z}_k; \mathbf{z}_j)
\end{equation}
We use the scikit-learn \citep{pedregosa2011scikit} implementation, which provides a robust, well-tested implementation of the KSG estimator. We use $k=3$ nearest neighbors as the default, and we leave the tuning of this parameter for future work.

\subsection{The TD2C Hypothesis}\label{sec:contribution-hypothesis}
\begin{figure}[ht]
    \centering
    \resizebox{\textwidth}{!}{    \begin{minipage}{1.3\linewidth}
    \begin{subfigure}[b]{0.32\textwidth}
        \centering
         
        \begin{tikzpicture}[
            node distance=0.5cm,
            greenroundnode/.style={circle, draw=black, fill=green!20, minimum size=6mm},
            grayroundnode/.style={circle, draw=black, fill=gray!20, minimum size=6mm},
            roundnode/.style={circle, draw=black, fill=none, minimum size=6mm},
            ]
            \node[greenroundnode]    (zi)  {$\mathbf{z}_i^{(t)}$};
            \node[roundnode]         (c2i) [above=of zi] {$\mathbf{z}_{i}^{(t-1)}$};
            \node[roundnode]         (e1i) [below=of zi] {$\mathbf{z}_{i}^{(t+1)}$};
            \node[roundnode]         (e2i) [below=of e1i] {$\mathbf{z}_{i}^{(t+2)}$};
            
            \node[greenroundnode]    (zj)  [right=of e1i] {$\mathbf{z}_j^{(t+1)}$};
            \node[roundnode]         (c1j) [above=of zj] {$\mathbf{z}_{j}^{(t)}$};  
            \node[roundnode]         (c2j) [above=of c1j] {$\mathbf{z}_{j}^{(t-1)}$};  
            \node[roundnode]         (e1j) [below=of zj] {$\mathbf{z}_{j}^{(t+2)}$};
            
            \draw[->, thick] (zi) -- (zj);
            \draw[->] (c2i) -- (c1j);
            \draw[->] (e1i) -- (e1j);
            \draw[->] (c2i) -- (zi) ;
            \draw[->] (c1j) -- (zj) ;
            \draw[->] (c2j) -- (c1j) ;
            \draw[->] (zi) -- (e1i);
            \draw[->] (e1i) -- (e2i) ;
            \draw[->] (zj) -- (e1j);
        \end{tikzpicture}
        \caption{$\left\{
            \begin{aligned} 
                \mathbf{z}_j^{(t+1)} &= f_{a}(\mathbf{z}_j^{(t)}, \mathbf{z}_i^{(t)}) + \boldsymbol{\epsilon}_j^{(t)} \\
                \mathbf{z}_i^{(t+1)} &= g_{a}(\mathbf{z}_i^{(t)}) + \boldsymbol{\epsilon}_i^{(t)}
            \end{aligned}
          \right.$}
        \label{fig:sub1}
    \end{subfigure}
    \hfill
    \begin{subfigure}[b]{0.32\textwidth}
        \centering
        \begin{tikzpicture}[
            node distance=0.5cm,
            greenroundnode/.style={circle, draw=black, fill=green!20, minimum size=6mm},
            grayroundnode/.style={circle, draw=black, fill=gray!20, minimum size=6mm},
            roundnode/.style={circle, draw=black, fill=none, minimum size=6mm},
            ]
            \node[greenroundnode]    (zi)  {$\mathbf{z}_i^{(t)}$};
            \node[roundnode]         (c2i) [above=of zi] {$\mathbf{z}_{i}^{(t-1)}$};
            \node[roundnode]         (e1i) [below=of zi] {$\mathbf{z}_{i}^{(t+1)}$};
            \node[roundnode]         (e2i) [below=of e1i] {$\mathbf{z}_{i}^{(t+2)}$};
            
            \node[greenroundnode]    (zj)  [right=of e1i] {$\mathbf{z}_j^{(t+1)}$};
            \node[roundnode]         (c1j) [above=of zj] {$\mathbf{z}_{j}^{(t)}$};  
            \node[roundnode]         (c2j) [above=of c1j] {$\mathbf{z}_{j}^{(t-1)}$};  
            \node[roundnode]         (e1j) [below=of zj] {$\mathbf{z}_{j}^{(t+2)}$};
            
            \draw[->, thick] (zi) -- (zj);
            \draw[->] (c2i) to[out=180, in=180] (e1i);
            \draw[->] (zi) to[out=180, in=180] (e2i);
            \draw[->] (c1j) to[out=0, in=0] (e1j);
            \draw[->] (c2j) to[out=0, in=0] (zj);
            \draw[->] (c2i) -- (c1j);
            \draw[->] (e1i) -- (e1j);
            \draw[->] (c2i) -- (zi) ;
            \draw[->] (c1j) -- (zj) ;
            \draw[->] (c2j) -- (c1j) ;
            \draw[->] (zi) -- (e1i);
            \draw[->] (e1i) -- (e2i) ;
            \draw[->] (zj) -- (e1j);
        \end{tikzpicture}
        \caption{$\left\{
            \begin{aligned} 
                \mathbf{z}_j^{(t+1)} &= f_{b}(\mathbf{z}_j^{(t)}, \mathbf{z}_i^{(t)}, \mathbf{z}_j^{(t-1)}) + \boldsymbol{\epsilon}_j^{(t)} \\
                \mathbf{z}_i^{(t+1)} &= g_{b}(\mathbf{z}_i^{(t)}, \mathbf{z}_i^{(t-1)}) + \boldsymbol{\epsilon}_i^{(t)}
            \end{aligned}
          \right.$}
        \label{fig:sub2}
    \end{subfigure}
    \hfill
    \begin{subfigure}[b]{0.32\textwidth}
        \centering
        \begin{tikzpicture}[
            node distance=0.5cm,
            greenroundnode/.style={circle, draw=black, fill=green!20, minimum size=6mm},
            grayroundnode/.style={circle, draw=black, fill=gray!20, minimum size=6mm},
            roundnode/.style={circle, draw=black, fill=none, minimum size=6mm},
            ]
            \node[greenroundnode]    (zi)   {$\mathbf{z}_i^{(t)}$};
            \node[roundnode]         (c2i)  [above=of zi] {$\mathbf{z}_{i}^{(t-1)}$};
            \node[roundnode]         (e1i)  [below=of zi] {$\mathbf{z}_{i}^{(t+1)}$};
            \node[roundnode]         (e2i)  [below=of e1i] {$\mathbf{z}_{i}^{(t+2)}$};
            
            \node[greenroundnode]    (zj)   [right=of e1i] {$\mathbf{z}_j^{(t+1)}$};
            \node[roundnode]         (c1j)  [above=of zj] {$\mathbf{z}_{j}^{(t)}$};  
            \node[roundnode]         (c2j)  [above=of c1j] {$\mathbf{z}_{j}^{(t-1)}$};  
            \node[roundnode]         (e1j)  [below=of zj] {$\mathbf{z}_{j}^{(t+2)}$};

            \node[grayroundnode]     (zk1)  [left=of e1i] {$\mathbf{L}^{(t+1)}$};
            \node[grayroundnode]     (zk)   [above=of zk1] {$\mathbf{L}^{(t)}$};
            \node[grayroundnode]     (zk_1) [above=of zk] {$\mathbf{L}^{(t-1)}$};
            \node[grayroundnode]     (zk2)  [below=of zk1] {$\mathbf{L}^{(t+2)}$};
            
            \draw[->, thick] (zi) -- (zj);
            \draw[->] (zk_1) -- (zi);
            \draw[->] (zk) -- (e1i);
            \draw[->] (zk1) -- (e2i);
            \draw[->] (zk_1) -- (c1j);
            \draw[->] (zk) -- (zj);
            \draw[->] (zk1) -- (e1j);
            \draw[->] (c2i) -- (c1j);
            \draw[->] (e1i) -- (e1j);
            \draw[->] (c2i) -- (zi) ;
            \draw[->] (c1j) -- (zj) ;
            \draw[->] (c2j) -- (c1j) ;
            \draw[->] (zk_1) -- (zk) ;
            \draw[->] (zk) -- (zk1) ;
            \draw[->] (zk1) -- (zk2) ;
            \draw[->] (zi) -- (e1i);
            \draw[->] (e1i) -- (e2i) ;
            \draw[->] (zj) -- (e1j);
        \end{tikzpicture}
        \caption{$\left\{
            \begin{aligned} 
                \mathbf{z}_j^{(t+1)} &= f_{c}(\mathbf{z}_j^{(t)}, \mathbf{z}_i^{(t)}, \mathbf{L}^{(t)}) + \boldsymbol{\epsilon}_j^{(t)} \\
                \mathbf{z}_i^{(t+1)} &= g_{c}(\mathbf{z}_i^{(t)}, \mathbf{L}^{(t)}) + \boldsymbol{\epsilon}_i^{(t)} 
            \end{aligned}
          \right.$}
        \label{fig:sub3}
       \end{subfigure}
        \end{minipage}    }
    \caption{Three causal scenarios illustrating the core hypothesis: while the asymmetry of the link $\mathbf{z}_i \rightarrow \mathbf{z}_j$ is obvious in the simple case (\subref{fig:sub1}), it becomes a more subtle, quantitative imbalance in the number of open paths in complex cases (\subref{fig:sub2}, \subref{fig:sub3}), as detailed in Table \ref{tab:paths-lenghts}. We hypothesize that this quantitative asymmetry provides a learnable signature of causality.}
    \label{fig:whole-ts}
\end{figure}
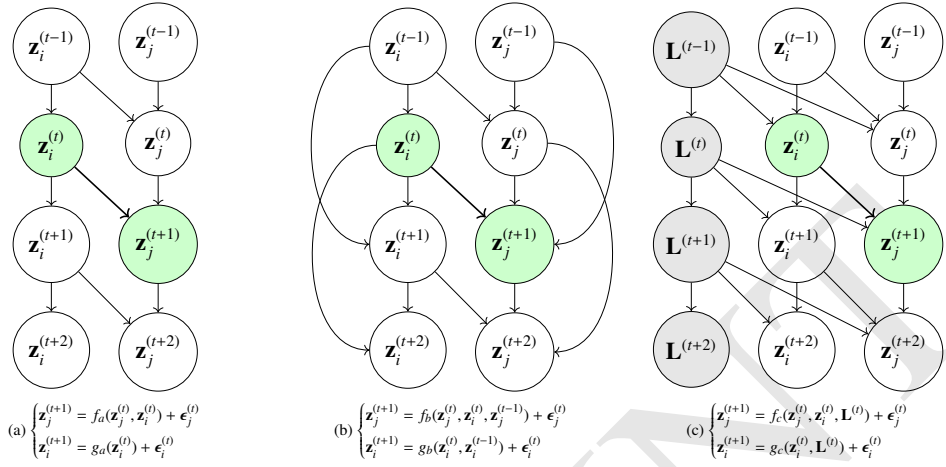

Our approach is founded on the hypothesis that a causal link, such as $\mathbf{z}_i^{(t)} \rightarrow \mathbf{z}_j^{(t+1)}$, creates a fundamental asymmetry in the flow of information through the temporal graph. Even when simple independence tests fail, a quantitative asymmetry persists. We posit that this underlying asymmetry manifests as systematic, learnable patterns in the distribution of non-zero mutual information terms. To illustrate this, we consider three scenarios of increasing complexity, depicted in Figure \ref{fig:whole-ts}.

\paragraph{Scenario 1: Simple Chain} A simple causal chain where $\mathbf{z}_i$ causes $\mathbf{z}_j$ (Figure \ref{fig:sub1}):
\begin{equation}
\left\{
 \begin{aligned}
    & \mathbf{z}_j^{(t+1)}=f_{a}(\mathbf{z}_j^{(t)}, \mathbf{z}_i^{(t)})+\boldsymbol{\epsilon}_j^{(t)} \\
    & \mathbf{z}_i^{(t+1)}=g_{a}(\mathbf{z}_i^{(t)})+\boldsymbol{\epsilon}_i^{(t)}
 \end{aligned} \right. \end{equation}

\paragraph{Scenario 2: Added Complexity} A longer-lag self-influence is added (Figure \ref{fig:sub2}):
\begin{equation}
\left\{
\begin{aligned}
    & \mathbf{z}_j^{(t+1)}=f_{b}(\mathbf{z}_j^{(t)},\mathbf{z}_i^{(t)})+\boldsymbol{\epsilon}_j^{(t)} \\
    & \mathbf{z}_i^{(t+1)}=g_{b}(\mathbf{z}_i^{(t)},\mathbf{z}_i^{(t-1)})+\boldsymbol{\epsilon}_i^{(t)}
 \end{aligned} \right. \end{equation}

\paragraph{Scenario 3: Confounding Variable} A confounder $\mathbf{L}$ influences both $\mathbf{z}_i$ and $\mathbf{z}_j$ (Figure \ref{fig:sub3}):
\begin{equation}
\left\{
 \begin{aligned}
    & \mathbf{z}_j^{(t+1)}=f_{c}(\mathbf{z}_j^{(t)},\mathbf{z}_i^{(t)}, \mathbf{L}^{(t)})+\boldsymbol{\epsilon}_j^{(t)} \\
    & \mathbf{z}_i^{(t+1)}=g_{c}(\mathbf{z}_i^{(t)},\mathbf{L}^{(t)})+\boldsymbol{\epsilon}_i^{(t)} \\
    & \mathbf{L}^{(t+1)}=h_{c}(\mathbf{L}^{(t)}) + \boldsymbol{\epsilon}_L^{(t)}
 \end{aligned} \right. \end{equation}

We analyzed the number of open d-separation paths for a Transfer Entropy-like test, $I(\mathbf{z}_i^{(t-k)}; \mathbf{z}_j^{(t)} \mid \mathbf{z}_j^{(t-k)})$, which isolates the unique information from the cause. We compared the number of open paths for this "Forward" test against its "Backward" counterpart, $I(\mathbf{z}_j^{(t-k)}; \mathbf{z}_i^{(t)} \mid \mathbf{z}_i^{(t-k)})$, as a function of lag $k$. To ensure comprehensive path enumeration, we extended our temporal DAGs across 20 time points (from $t = -10$ to $t = 10$), allowing us to increase the number of potential open paths significantly and further highlight the asymmetry. The results (Table \ref{tab:paths-lenghts}) support our hypothesis: even when both directions have open paths, a quantitative asymmetry ($\Delta$) persists and grows with complexity, providing a learnable signal. The extensive temporal unrolling ensures that all relevant causal pathways within our maximum lag horizon are properly represented in the d-separation analysis.

\begin{table*}[ht]
\centering
\small
\caption{Number of open information paths for different causal scenarios and time lags ($k$). The analysis was performed on a DAG unrolled for 20 time steps. The asymmetry between the forward and backward directions, measured by the difference ($\Delta$), persists and often grows with the complexity of the causal graph. We hypothesize that this provides a robust, learnable signature for causal directionality.}
\label{tab:paths-lenghts}
\begin{tabular}{@{}lccccccccc@{}}
\toprule
& \multicolumn{9}{c}{\textbf{Time Lag (k)}} \\
\cmidrule(l){2-10}
\textbf{Test Direction} & \textbf{1} & \textbf{2} & \textbf{3} & \textbf{4} & \textbf{5} & \textbf{6} & \textbf{7} & \textbf{8} & \textbf{9} \\
\midrule
\multicolumn{10}{@{}l}{\textbf{Scenario 1: Simple Case (Fig. \ref{fig:sub1})}} \\
\quad Forward ($\rightarrow$) & 1 & 2 & 3 & 4 & 5 & 6 & 7 & 8 & 9 \\
\quad Backward ($\leftarrow$) & 0 & 0 & 0 & 0 & 0 & 0 & 0 & 0 & 0 \\
\quad \textbf{Difference ($\Delta$)} & \textbf{1} & \textbf{2} & \textbf{3} & \textbf{4} & \textbf{5} & \textbf{6} & \textbf{7} & \textbf{8} & \textbf{9} \\
\midrule
\multicolumn{10}{@{}l}{\textbf{Scenario 2: Added Complexity (Fig. \ref{fig:sub2})}} \\
\quad Forward ($\rightarrow$) & 370 & 228 & 324 & 320 & 374 & 408 & 469 & 529 & 615 \\
\quad Backward ($\leftarrow$) & 312 & 153 & 227 & 183 & 194 & 176 & 152 & 125 & 71 \\
\quad \textbf{Difference ($\Delta$)} & \textbf{58} & \textbf{75} & \textbf{97} & \textbf{137} & \textbf{180} & \textbf{232} & \textbf{317} & \textbf{404} & \textbf{544} \\
\midrule
\multicolumn{10}{@{}l}{\textbf{Scenario 3: Latent Confounding Variable (Fig. \ref{fig:sub3})}} \\
\quad Forward ($\rightarrow$) & 1186 & 880 & 1138 & 1123 & 1187 & 1209 & 1234 & 1245 & 1271 \\
\quad Backward ($\leftarrow$) & 898 & 502 & 701 & 580 & 596 & 534 & 479 & 403 & 290 \\
\quad \textbf{Difference ($\Delta$)} & \textbf{288} & \textbf{378} & \textbf{437} & \textbf{543} & \textbf{591} & \textbf{675} & \textbf{755} & \textbf{842} & \textbf{981} \\
\bottomrule
\multicolumn{10}{l}{\makecell[l]{Note: Forward ($\rightarrow$) corresponds to paths for $I(\mathbf{z}_i^{(t-k)}; \mathbf{z}_j^{(t)} \mid \mathbf{z}_j^{(t-k)})$. \\ Backward ($\leftarrow$) corresponds to paths for $I(\mathbf{z}_j^{(t-k)}; \mathbf{z}_i^{(t)} \mid \mathbf{z}_i^{(t-k)})$.}} \\
\end{tabular}
\end{table*}

\subsection{From Hypothesis to Descriptors}\label{sec:contribution-operationalize}

Motivated by the quantitative asymmetry demonstrated in our path analysis, we operationalize this hypothesis by introducing a family of descriptors based on a generalized form of Transfer Entropy (TE). Our approach systematically evaluates how the information from the cause's immediate past, $\mathbf{z}_i^{(t-1)}$, transfers to the effect's present, $\mathbf{z}_j^{(t)}$, under varying temporal contexts of the effect. By fixing the cause at its most recent lag, we focus on the most direct potential influence, consistent with the standard TE formulation. The generalization comes from varying the conditioning history of the effect, which allows us to probe the persistence of this information flow.

For each conditioning lag $k$ in a predefined range (in our experiments, $k_{min}=1$ to $k_{max}=15$), we compute:
\begin{align}
\text{TE}_{\text{fwd}}^{(k)} &= I(\mathbf{z}_i^{(t-1)}; \mathbf{z}_j^{(t)} \mid \mathbf{z}_j^{(t-k)}) \label{eq:te_fwd}\\
\text{TE}_{\text{bwd}}^{(k)} &= I(\mathbf{z}_j^{(t-1)}; \mathbf{z}_i^{(t)} \mid \mathbf{z}_i^{(t-k)}) \label{eq:te_bwd}
\end{align}
From these, we derive several features. The primary feature, our generalized TE asymmetry, averages the directional bias across all lags:
\begin{equation}
\text{TE}_{\text{asy}} = \mathbb{E}_{k \in [k_{min}, k_{max}]}\left[\text{TE}_{\text{fwd}}^{(k)} - \text{TE}_{\text{bwd}}^{(k)}\right] \label{eq:te_asymmetry}
\end{equation}
We also include the standard TE measures (for $k=1$) and their difference to capture direct, short-term relationships:
\begin{align}
\text{TE}_{\text{fwd}}^{(1)} &= I(\mathbf{z}_i^{(t-1)}; \mathbf{z}_j^{(t)} \mid \mathbf{z}_j^{(t-1)}) \label{eq:te_fwd_1}\\
\text{TE}_{\text{bwd}}^{(1)} &= I(\mathbf{z}_j^{(t-1)}; \mathbf{z}_i^{(t)} \mid \mathbf{z}_i^{(t-1)}) \label{eq:te_bwd_1}\\
\Delta\text{TE}^{(1)} &=\text{TE}_{\text{fwd}}^{(1)} - \text{TE}_{\text{bwd}}^{(1)} \label{eq:te_diff_1}
\end{align}
This multi-lag approach provides robustness against the choice of conditioning window while capturing both broad temporal patterns and specific short-term causal signals.

It is important to note the subtle distinction between this operationalization and the path-counting experiment in Section \ref{sec:contribution-hypothesis}, where the cause $\mathbf{z}_i$ was also lagged by $k$. Our implemented descriptor prioritizes the direct influence from the immediate past ($\mathbf{z}_i^{(t-1)}$) as a robust and empirically effective starting point. We leave the exploration of the fully lagged descriptor form, $I(\mathbf{z}_i^{(t-k)}; \mathbf{z}_j^{(t)} \mid \mathbf{z}_j^{(t-k)})$, as well as the systematic tuning of the lag range $[k_{min}, k_{max}]$, as promising directions for future work.

\subsection{Other Novel Descriptor Families}\label{sec:contribution-novel}
To create a comprehensive model, we introduce three additional descriptor families.

\subsubsection{Error-Based Descriptors}\label{sec:contribution-novel-error}

This family of descriptors draws inspiration from the principles of error-based causal discovery, particularly from Additive Noise Models (ANMs) \citep{hoyer2008nonlinear}. The core idea is that a true causal relationship imposes a specific structure on the residuals of a predictive model, creating asymmetries that are not present in the reverse, non-causal direction. We leverage this principle to introduce two complementary error-based features.

First, we compute the partial correlation to isolate the direct linear association between $\mathbf{z}_i^{(t)}$ and $\mathbf{z}_j^{(t)}$ by removing the confounding effects of their temporal neighbors. To do so, we define a combined conditioning set, $\mathbf{S}_{ij}^{(t)} = \mathbf{M}_i^{(t)} \cup \mathbf{M}_j^{(t)}$, containing all variables in both temporal Markov Blankets. We then use two separate regression models, $f_i$ and $f_j$, to predict each variable from this common set. The partial correlation is the correlation between the resulting residuals, $\boldsymbol{\epsilon}_i$ and $\boldsymbol{\epsilon}_j$:
\begin{align}
\boldsymbol{\epsilon}_i &= \mathbf{z}_i^{(t)} - f_i(\mathbf{S}_{ij}^{(t)}) \\
\boldsymbol{\epsilon}_j &= \mathbf{z}_j^{(t)} - f_j(\mathbf{S}_{ij}^{(t)}) \\
\rho_{\text{partial}} &= \text{Corr}(\boldsymbol{\epsilon}_i, \boldsymbol{\epsilon}_j) \label{eq:partial_corr}
\end{align}
A non-zero value for this descriptor suggests a direct link that is not mediated by the immediate temporal neighbors included in the conditioning set.

Second, we compute the residual-input correlation to test for a different kind of asymmetry. This is based on the ANM principle that for the correct causal direction $\mathbf{z}_i \to \mathbf{z}_j$, the residual of a model predicting the effect should be independent of the cause. We test this by building a predictive model for the effect using the cause and the cause's own Markov Blanket, $\mathbf{M}_i^{(t)}$, as predictors. A non-zero correlation between the residual and the input cause suggests that the causal variable holds information beyond the model's capacity, an asymmetry not expected in the reverse, non-causal direction.
\begin{align}
\boldsymbol{\epsilon}_j &= \mathbf{z}_j^{(t)} - f_{\text{Ridge}}(\mathbf{z}_i^{(t)}, \mathbf{M}_i^{(t)}) \\
\rho_{\text{resid}} &= \text{Corr}(\boldsymbol{\epsilon}_j, \mathbf{z}_i^{(t)}) \label{eq:resid_corr}
\end{align}

\subsubsection{Higher-Order Moment Descriptors}\label{sec:contribution-novel-hoc}
This family captures non-Gaussianity and distributional asymmetries using statistical moments. We compute specific cross-cumulants:
\begin{equation}\label{eq:hoc}
\text{HOC}_{i,j} = \mathbb{E}\left[\left({(\mathbf{z}_i^{(t)} - \mu_i)}/{\sigma_i}\right)^i \left({(\mathbf{z}_j^{(t)} - \mu_j})/{\sigma_j}\right)^j\right]
\end{equation}
where $\mu$ and $\sigma$ are the mean and standard deviation. We compute $\text{HOC}_{3,1}$, $\text{HOC}_{1,2}$, $\text{HOC}_{2,1}$, $\text{HOC}_{1,3}$, as well as the univariate Kurtosis and Skewness for each variable.

\subsubsection{Linear Descriptors}\label{sec:contribution-novel-linear}
Linearity is the most fundamental type of statistical dependency. For a potential link from $\mathbf{z}_i$ to $\mathbf{z}_j$, a simple linear descriptor is the regression coefficient $b$ obtained from a linear model that predicts the target variable from the source variable and the target's Markov Blanket:

\begin{equation}
\mathbf{z}_j = b \cdot \mathbf{z}_i + \sum_{k} w_k \mathbf{m}_{j,k} + \boldsymbol{\epsilon}
\end{equation}

The coefficient $b$ quantifies the direct linear association between $\mathbf{z}_i$ and $\mathbf{z}_j$ after controlling for the other members of the Markov Blanket. This is computed for both the forward ($\mathbf{z}_i \rightarrow \mathbf{z}_j$) and backward ($\mathbf{z}_j \rightarrow \mathbf{z}_i$) directions, providing a simple, asymmetric measure of linear dependence.

\subsection{Final Descriptor Set}\label{sec:final-descriptor-set}

Our final model enhances the proven legacy descriptors from \citep{bontempi2020learning} with a new set of 16 features specifically designed to capture temporal and non-linear dynamics. This novel set is composed of: the four generalized Transfer Entropy measures (Equations~\ref{eq:te_asymmetry}-\ref{eq:te_diff_1}); the two error-based descriptors (Equations~\ref{eq:partial_corr} and \ref{eq:resid_corr}); the eight higher-order moment statistics derived from Equation~\ref{eq:hoc}; and the two linear regression coefficients. Together, these novel features and the retained legacy descriptors form the complete feature vector used by TD2C. A comprehensive list of all descriptors is provided in \ref{app:full_descriptors}, Tables \ref{tab:descriptor_formulas_math_p1} and ~\ref{tab:descriptor_formulas_math_p2}.

\subsection{From Descriptors to Causal Probabilities}\label{sec:td2c-procedure}
At its core, TD2C operationalizes the principle of causal asymmetry through a supervised learning pipeline. This pipeline systematically translates the complex statistical signatures surrounding a potential causal link into a single, interpretable probability. The entire process, from raw time series to a trained causal discovery model, is formalized in \ref{app:pipeline} (Algorithm~\ref{alg:pipeline}) and can be summarized in the following stages:

\begin{enumerate}
    \item \textbf{Data Reshaping and Pair Selection:} The first step is to transform each raw time series of shape $(T, N)$ into a static data matrix suitable for feature extraction. This is done by creating a lagged representation, resulting in a matrix of shape $(T-L) \times (N \cdot (L+1))$, where $L$ is the maximum considered lag. Each column in this matrix represents a specific variable at a specific time lag (e.g., $\mathbf{z}_i^{(t)}, \mathbf{z}_i^{(t-1)}$, etc.). Crucially, to maintain a focus on temporal causality and ensure computational tractability, we do not test all possible pairs of nodes in this flattened representation. Instead, we only consider candidate links from the past to the present, of the form $\mathbf{z}_i^{(t-\tau)} \to \mathbf{z}_j^{(t)}$, where the lag $\tau$ is between 1 and the maximum lag $L$.

    \item \textbf{Feature Extraction:} For every candidate link identified in the previous step, we define their temporal MBs based on the self-causality assumption. We then compute the full vector of descriptors (as defined in Section \ref{sec:contribution-novel} and \ref{sec:final-descriptor-set}) to quantify the statistical relationship and information flow.
    
    \item \textbf{Training:} Each generated feature vector is assigned a binary label—causal (1) or non-causal (0)—based on the ground-truth causal graph of its source time series. By aggregating these labeled vectors from a large collection of synthetic datasets, we create a rich training set to train a supervised classifier, which learns the complex, non-linear patterns that differentiate true causal links from spurious correlations.
    
    \item \textbf{Inference:} For a new, unlabeled time series, we apply the exact same reshaping and descriptor extraction pipeline. The pre-trained classifier then takes the feature vector for any candidate link as input and outputs the predicted probability that a causal relationship exists.
\end{enumerate}

\section{Experimental Setup}\label{sec:experiments}
This section presents our experimental setup. We first describe our synthetic data generation process using multivariate NAR processes (\ref{sec:synthetic-data-generation}), followed by our evaluation on established benchmarks including NetSim and DREAM3 datasets (\ref{sec:realistic-benchmark-data}). We compare against state-of-the-art causal discovery methods spanning multiple paradigms (\ref{sec:benchmark-setup}) using appropriate evaluation metrics for the class-imbalanced nature of causal discovery (\ref{sec:evaluation-metrics}). Finally, we detail our threshold selection procedure (\ref{sec:exp-threhsold}) and statistical testing framework (\ref{sec:statistical-tests-methodology}) to ensure robust performance comparisons.

\subsection{Synthetic Data Generation}\label{sec:synthetic-data-generation}
To build a robust training and testing foundation, we generated a rich synthetic dataset from our library of multivariate Nonlinear Autoregressive (NAR) processes inspired by \citep{bontempi2020learning}. Details are available in \ref{app:synthetic-data} in Table \ref{tab:time-series-generative-processes}. For these synthetic experiments, we focus exclusively on time series with $N=5$ variables. This choice is deliberate; while a larger $N$ would increase the number of candidate pairs from a single time series, our primary goal is to assess the model's ability to generalize across a wide variety of causal dynamics. We therefore prioritize generating numerous time series across many different underlying processes over simply increasing the dimensionality of a few. The scalability of our method to higher dimensions is evaluated separately using the realistic benchmark data in subsequent sections. \\

To ensure a fair comparison with methods that have specific error assumptions (e.g., VarLiNGAM), we generated data using three distinct noise distributions: Gaussian, Uniform, and Laplace. For each of the 9 NAR processes designated for training (Processes 1, 3, 5, 7, 9, 11, 13, 15, 19) and for each noise distribution, we generated 120 unique time series of 250 timesteps (3240 series, 243000 pairs). The classifier's performance is then evaluated on a separate set of even-indexed processes (Processes 2, 4, 6, 8, 10, 12, 14, 16, 18), from which we generated 40 unique time series of 250 timesteps for each noise distribution (1080 series, 81000 pairs). This odd/even split tests whether our method has learned fundamental characteristics of causality rather than merely overfitting to the training process dynamics. To ensure data quality, we implemented a stability check that discards any generated time series containing non-finite or divergent values beyond a threshold of $10^6$. The initial state for each series, required to bootstrap the autoregressive process, is drawn from a uniform distribution $U(-1, 1)$. A summary of the generation parameters can be found in Table \ref{tab:generation-parameters}. An example of a generated time series with $N=5$ and its corresponding DAG are shown in \ref{app:example-generated-time-series}, in Table \ref{tab:time_series_data_a} and Figure \ref{fig:example-dag-a}, respectively.

\begin{table}[ht]
\resizebox{\linewidth}{!}{
\begin{tabular}{|l|l|}
\hline
Parameter         & Description                                        \\ \hline
Generative Process $p$          & $p \in \{1,2,3,4,5,6,7,8,9,10,11,12,13,14,15,16,18,19\}$                    \\ \hline
Variables per Series $N$       & $N = 5$         \\ \hline
Observations per combination $O$       & $O=120$ for training $O=40$ for testing         \\ \hline
Noise distributions   & Gaussian, Laplace, Uniform  \\ \hline
Interactions               & Dynamic, capturing linear and nonlinear relationships       \\ \hline
Lag Structures             & Complex, with temporal dependencies up to 4 time steps      \\ \hline
Max Parents set size $\eta_j^{max}$    & $\eta_j^{max} = 2$  \\ \hline
Parents set selection mechanism     & Stochastic selection, different for each variable                      \\ \hline
Initial State Distribution & Uniform random distribution between -1 and 1                \\ \hline
Evolution Over Time        & Each variable influenced by its parents set, $\eta_j$ (random)         \\ \hline
Timesteps per Series $L$       & $L=250$, to observe variable evolution and causal relationships \\ \hline
\end{tabular}
}
\caption{Summary of the parameters of our synthetic time series generation, adapted from \citep{bontempi2020learning}}
\label{tab:generation-parameters}
\end{table}

\subsection{Realistic Benchmark Data}\label{sec:realistic-benchmark-data}
To evaluate how well our synthetically-trained method generalizes to real-world scenarios, we utilize two widely recognized benchmark suites for testing:

\begin{itemize}
    \item NetSim: a network inference simulator \citep{smith2011network} that generates dynamical data from known biological network structures. We use datasets with 5 and 10 variables, referred to as NETSIM-5 and NETSIM-10. NETSIM-5 with 1,050 time series samples, each containing 5 variables observed over different temporal lengths (mostly 200 timesteps), and NETSIM-10 with 250 time series samples of 10 variables, where most sequences contain 200 timesteps with a few exception of 1000+ timesteps.
    \item DREAM3 In Silico Challenge: a systems biology benchmark \citep{prill2010towards} providing simulated gene regulatory network data. We use datasets with 10 and 50 variables, referred to as DREAM3-10 and DREAM3-50. DREAM3-10 contains 5 time series of 10 variables each observed over 84 timesteps, and DREAM3-50 features 5 time series of 50 variables each spanning 483 timesteps.
\end{itemize}

Adapting these benchmarks to our time-series framework required a set of hypotheses to construct the ground-truth temporal DAGs from their provided static network structures, despite their temporal observational data. Specifically, to create a time-unrolled DAG, we assume that every variable's state at time $t$ is directly caused by its own state at time $t-1$. This adds an autoregressive edge for every variable $i$ in the system. This is a standard and reasonable assumption for dynamical systems data. For a known static causal link from variable $i$ to $j$, where the precise time lag is unknown, we assume a default lag of one time step. This translates the static edge into a temporal edge. To ensure a fair comparison, a maximum lag of $L=3$ was used for all applicable methods, including TD2C, PCMCI, and the VAR-based approaches. This tests each algorithm's ability to discover causal links up to this common temporal depth.\\

While these assumptions simplify the true, and often unknown, temporal dynamics of the original systems, they allow us to create a consistent ground truth for evaluation. It is important to note that any potential mismatch with the real underlying dynamics affects all benchmarked methods equally. Therefore, this approach does not invalidate the relative comparison between methods but rather tests their ability to discover causal links under this common, plausible set of temporal assumptions. These datasets serve as the ultimate test for generalization from our synthetic training environment to more realistic and complex systems.

\subsection{Benchmark Methods and Implementation}\label{sec:benchmark-setup}

A comprehensive comparison against multiple state-of-the-art causal discovery methodologies is performed. Each selected method exemplifies a distinct category within the spectrum of causal inference approaches. As a simple yet powerful baseline, we include a direct coefficient-based VAR method. In this approach, a causal link is inferred if the corresponding coefficient in the fitted VAR model is both statistically significant (p-value $<0.05$ ) and has a magnitude greater than 0.1. From the time-series causality family, we include two complementary approaches: the Pairwise Granger Causality test implementation available in the \texttt{statsmodels} package for Python~\citep{seabold2010statsmodels}, which provides a comprehensive suite for econometric and statistical modeling, and Multivariate Granger Causality (MVGC) implemented through Vector Autoregression (VAR) models. The MVGC approach extends pairwise analysis by considering the full multivariate context when testing for Granger causality, using F-tests on VAR model coefficients to determine causal relationships while accounting for the influence of all variables in the system. From the constraint-based family, we select the PCMCI algorithm developed by \citep{runge2019detecting}, which is implemented in the Tigramite Python package. The VarLiNGAM method, as proposed by \citep{hyvarinen2010estimation}, is our choice for the noise-based category. It is implemented in Python via the LiNGAM library \citep{ikeuchi2023python}. Finally, for the score-based category, we incorporate DYNOTEARS, a method introduced \citep{pamfil2020dynotears} and implemented in the CausalNex Python library~\citep{beaumont_CausalNex_2021}. A summary of all competitors can be found in \ref{app:competitors}, Table \ref{tab:competitors}.\\

It is important to note that each method has its own underlying assumptions, which might not always be respected in practical scenarios. For example, VarLiNGAM assumes non-Gaussian errors, reason for which we include Laplace and Uniform distributed noise. For PCMCI, we employ both the ParCorr independence test and the GPDC (Gaussian Process Distance Correlation) test to capture linear and nonlinear dependencies respectively. We were unable to use the k-nearest neighbor conditional mutual information independence test due to its long execution time, which made it computationally infeasible for our experimental setup. The MVGC implementation assumes linear relationships through its VAR model foundation and may miss nonlinear causal dependencies.

\subsection{Evaluation Protocol}\label{sec:evaluation-metrics}

Given the unbalanced nature of the problem at hand (i.e. there is significant more non-causal couples than causal couples) we adopt a BalancedRandomForest Classifier from the imblearn python package \citep{JMLR:v18:16-365}. We leave the tuning of this classifier for future work. The subsequent analysis is conducted within the framework of binary classification, focusing on the minority class. In this context, we define True Positives (TP) as correctly identified causal couples, False Positives (FP) as non-causal couples incorrectly classified as causal, True Negatives (TN) as correctly identified non-causal couples, and False Negatives (FN) as causal couples incorrectly classified as non-causal. We then define the following performance metrics:

\begin{equation}\label{eq:precision}
\text{Precision}=\frac{\mathrm{TP}}{\mathrm{TP}+\mathrm{FP}}
\end{equation}
\begin{equation}\label{eq:recall}
\text{Recall}=\frac{\mathrm{TP}}{\mathrm{TP}+\mathrm{FN}}
\end{equation}
\begin{equation}\label{eq:f1}
\text{F1-Score}=\frac{2 \cdot \text{Precision} \cdot \text{Recall}}{\text{Precision} + \text{Recall}}
\end{equation}
\begin{equation}\label{eq:accuracy}
\text{Accuracy}=\frac{\mathrm{TP}+\mathrm{TN}}{\mathrm{TP}+\mathrm{TN}+\mathrm{FP}+\mathrm{FN}}
\end{equation}
\begin{equation}\label{eq:balanced_accuracy}
\text{Balanced Accuracy}=\frac{1}{2}\left(\frac{\mathrm{TP}}{\mathrm{TP}+\mathrm{FN}}+\frac{\mathrm{TN}}{\mathrm{TN}+\mathrm{FP}}\right)
\end{equation}
Precision (Equation~\ref{eq:precision}) calculates the proportion of actual positives among the predicted positives. Recall (Equation~\ref{eq:recall}), or the True Positive Rate (TPR), quantifies the proportion of actual positives that were correctly identified. The F1-Score (Equation~\ref{eq:f1}) provides the harmonic mean of precision and recall, offering a single metric that balances both measures. Accuracy (Equation~\ref{eq:accuracy}) measures the overall proportion of correct predictions, while Balanced Accuracy (Equation~\ref{eq:balanced_accuracy}) adjusts for class imbalance by averaging sensitivity (recall) and specificity, making it particularly suitable for our unbalanced dataset. We also define the False Positive Rate (FPR) as the proportion of actual negatives that are incorrectly classified as positives.

\subsection{Threshold selection}\label{sec:exp-threhsold}
The classifier component of our TD2C method outputs a continuous probability score for each potential causal link. Ideally, we would evaluate performance using Receiver Operating Characteristic (ROC) curves, which plot TPR against FPR across all possible thresholds, or Precision-Recall (PR) curves, which plot Precision against Recall. However, to enable a fair comparison with benchmark methods that produce only binary outputs, a single decision threshold ($\tau$) must be established. A default threshold of 0.5 is often suboptimal, particularly given the severe class imbalance inherent to causal discovery. We therefore developed a robust procedure to select a data-informed threshold based solely on the training data. To ensure the resulting threshold generalizes to unseen causal structures, we employed a Leave-One-Process-Out Cross-Validation strategy. In this 9-fold procedure, we repeatedly trained the classifier on data from 8 causal processes and determined an optimal threshold by evaluating its performance on the 9th, entirely held-out process. \\

We conducted a sensitivity analysis within each fold, comparing four distinct criteria: maximizing the F1-Score \eqref{eq:f1}, the harmonic mean of Precision and Recall, the Precision-Recall Break-Even Point, where Precision and Recall values are numerically closest, maximizing Youden's J Statistic \citep{youden1950index}, the difference between TPR and FPR, and minimizing $D_{ROC(0,1)}$, the distance to the "perfect classifier" corner (0,1) in the ROC curve \citep{fawcett2006introduction}. The final metric was chosen based on two pre-defined principles: domain relevance (favoring recall to minimize missed discoveries) and statistical stability (the lowest standard deviation across folds).

\subsection{Statistical Tests}\label{sec:statistical-tests-methodology}

For a robust statistical evaluation of the different causal discovery algorithms, we adopt the framework proposed by \citep{demvsar2006statistical}. We first conduct a Friedman test to assess the global null hypothesis that all methods perform equally across the test datasets. If this test is significant ($p < 0.05$), we proceed with post-hoc pairwise Wilcoxon signed-rank tests to compare each pair of methods directly, with p-values corrected using the Holm-Bonferroni procedure to control the family-wise error rate. The final average ranks and statistically significant differences are visualized using Critical Difference (CD) diagrams.

A critical aspect of this framework is the scope of its application. These tests will be applied exclusively to the 1080 runs from our held-out synthetic test set. The justification for this choice is the fundamental requirement of statistical independence for the validity of the Friedman and Wilcoxon tests. Our synthetic dataset's generative process, which involves the random initialization of observations and the random selection of parent sets for each variable, ensures that each test run can be treated as an independent and identically distributed (IID) observation. In contrast, the real-world benchmark datasets (e.g., DREAM3, NetSim) have fixed underlying causal graphs. Their various time series observations therefore cannot be treated as independent samples for comparing algorithm performance, making the application of these statistical tests to them statistically unsound.

\section{Results}\label{sec:results}
This section presents the empirical findings from our comprehensive evaluation of TD2C against state-of-the-art causal discovery methods. We report our selected threshold (Section~\ref{sec:threshold-selection-for-td2c}), followed by the performance outcomes on held-out synthetic data, demonstrating generalization to unseen causal dynamics (Section~\ref{sec:unseen-dynamics}), and on established realistic benchmarks, revealing zero-shot capabilities on real-world scenarios (Section~\ref{sec:zero-shot-generalization}). We present rigorous statistical comparisons to establish the significance of performance differences (Section~\ref{sec:results-statistical-test}), followed by computational performance analysis and interpretability insights through feature importance examination (Section~\ref{sec:analysis-computational-performance}).

\subsection{Threshold Selection for TD2C}\label{sec:threshold-selection-for-td2c}

The Leave-One-Process-Out cross-validation described in our methodology yielded the results summarized in Table~\ref{tab:threshold_metrics}. The analysis confirms that the choice of optimization metric significantly impacts the resulting threshold, with average values ranging from 0.309 to 0.618. Following the selection criteria outlined in our methodology, namely domain relevance and statistical stability, the $D_{ROC(0,1)}$ metric was identified as the most suitable. As shown in the table, it produced both the lowest average threshold and the lowest standard deviation. Based on this rigorous, data-driven procedure, we selected a final, robust threshold of $\tau = 0.309$ for TD2C. This threshold was used for all subsequent performance evaluations and comparisons against benchmark methods.

\begin{table}[!ht]
\centering
\caption{Comparison of optimization metrics for determining the TD2C decision threshold. Results are averaged across a 9-fold Leave-One-Process-Out cross-validation on the training set. The $D_{ROC(0,1)}$ metric was selected (\textbf{bold}) for its combination of high stability (lowest Std. Dev.) and its tendency to favor recall (lowest Avg. Threshold).}
\label{tab:threshold_metrics}
\begin{tabular}{@{}lcc@{}}
\toprule
\textbf{Optimization Metric} & \textbf{Average Threshold} & \textbf{Std. Deviation} \\ \midrule
Maximize F1-Score            & 0.478                      & 0.263                   \\
Precision-Recall Break-Even  & 0.618                      & 0.275                   \\
Maximize Youden's J          & 0.382                      & 0.268                   \\
Minimize $D_{ROC(0,1)}$ & \textbf{0.309}          & \textbf{0.206}         \\ \bottomrule
\end{tabular}
\end{table}

\subsection{Performance on Synthetic Data: Generalization to Unseen Dynamics}\label{sec:unseen-dynamics}
Our primary test of generalization involves evaluating TD2C on the held-out synthetic dataset. As detailed in Section~\ref{sec:experiments}, this dataset includes data generated with Gaussian, Uniform, and Laplace noise distributions from nine distinct NAR processes not seen during training. The performance of TD2C and competing methods, aggregated across all 1080 test runs, is summarized in Table~\ref{tab:overall_results}. The results clearly demonstrate the superior generalization capability of our proposed method. TD2C achieves the highest Balanced Accuracy (0.8218) and a substantially higher F1-Score (0.6306) than all other evaluated methods. This is a significant improvement over the next best methods in terms of F1-Score, such as PCMCI (0.4959) and MVGC (0.4886). While some methods like PCMCI-GPDC achieve a slightly higher raw Accuracy (0.8704 vs. 0.8533 for TD2C), their considerably lower Balanced Accuracy and F1-Scores indicate a bias towards the majority class (i.e., predicting no causal link) and a weaker ability to correctly identify true causal relationships. In contrast, TD2C provides the best balance between precision and recall, as reflected in its leading F1-Score. Other traditional methods, such as Granger causality and DYNOTEARS, struggle to handle the complex, nonlinear dynamics, resulting in very low performance across all relevant metrics. \\

\begin{table}[!ht]
\centering
\caption{Overall performance on the held-out synthetic dataset. Scores are macro-averaged across all 1080 runs (9 processes $\times$ 3 noise types $\times$ 40 instances), presented as mean $\pm$ standard deviation. TD2C achieves the best F1-Score and Balanced Accuracy, indicating superior and more reliable causal discovery. Best performance for each metric is highlighted in \textbf{bold}.}
\label{tab:overall_results}
\resizebox{\textwidth}{!}{\begin{tabular}{lccccc}
\toprule
\textbf{Method} & \textbf{Accuracy} & \textbf{Balanced Accuracy} & \textbf{F1-Score} & \textbf{Precision} & \textbf{Recall} \\
\midrule
TD2C         & 0.8533 $\pm$ 0.0972 & \textbf{0.8218 $\pm$ 0.1344} & \textbf{0.6306 $\pm$ 0.2126} & \textbf{0.5637 $\pm$ 0.2368} & \textbf{0.7708 $\pm$ 0.2241} \\
DYNOTEARS    & 0.8003 $\pm$ 0.1499 & 0.5325 $\pm$ 0.0739          & 0.1043 $\pm$ 0.1945          & 0.1805 $\pm$ 0.3652          & 0.1342 $\pm$ 0.2433          \\
Granger      & 0.7155 $\pm$ 0.1277 & 0.4487 $\pm$ 0.0586          & 0.0607 $\pm$ 0.0829          & 0.0674 $\pm$ 0.1043          & 0.0686 $\pm$ 0.0995          \\
MVGC         & 0.8644 $\pm$ 0.0733 & 0.7409 $\pm$ 0.1940          & 0.4886 $\pm$ 0.3243          & 0.4850 $\pm$ 0.2999          & 0.5638 $\pm$ 0.4055          \\
PCMCI        & 0.8674 $\pm$ 0.0722 & 0.7343 $\pm$ 0.1999          & 0.4959 $\pm$ 0.3301          & 0.4976 $\pm$ 0.2988          & 0.5414 $\pm$ 0.3981          \\
PCMCI-GPDC   & \textbf{0.8704 $\pm$ 0.0671} & 0.7232 $\pm$ 0.1926          & 0.4841 $\pm$ 0.3256          & 0.5132 $\pm$ 0.3029          & 0.5075 $\pm$ 0.3844          \\
VAR          & 0.8371 $\pm$ 0.0608 & 0.5365 $\pm$ 0.0557          & 0.1444 $\pm$ 0.1529          & 0.3351 $\pm$ 0.3522          & 0.1008 $\pm$ 0.1162          \\
VARLiNGAM    & 0.7928 $\pm$ 0.1351 & 0.7115 $\pm$ 0.1917          & 0.4501 $\pm$ 0.2909          & 0.4114 $\pm$ 0.2613          & 0.5875 $\pm$ 0.3699          \\
\bottomrule
\end{tabular}
}
\end{table}

A notable feature of the results is the relatively high standard deviation for most metrics, particularly for the F1-Score. This variance is not an indicator of model instability but rather reflects the diverse and challenging nature of the test set. The dataset comprises multiple distinct NAR processes, each presenting unique challenges in terms of dynamics, lag structure, and signal-to-noise ratio. A method may perform exceptionally well on one type of process but struggle with another, leading to a wide distribution of scores when aggregated. The strong mean performance of TD2C, despite this inherent variance in task difficulty, underscores its robustness across a wider range of these challenging processes compared to its competitors. A more detailed breakdown of performance for each of the nine NAR processes is provided in Table~\ref{tab:combined_results_long} in \ref{app:extra-results-synthetic}. For instance, on processes like 12 and 18, TD2C is the top-performing method across every single metric, showcasing its effectiveness on certain classes of problems. Conversely, on Process 4, MVGC shows a competitive edge, highlighting that no single method is a panacea for all types of causal structures. Furthermore, the results for processes such as 6 and 8 show a general decrease in performance across all methods, indicating these generative processes are inherently more challenging. The key takeaway is that while other methods may excel on specific, narrow tasks, TD2C provides the most robust and high-performing solution across the diverse set of unseen dynamics, reinforcing the conclusions from the aggregated results in Table~\ref{tab:overall_results}.

\subsection{Performance on Realistic Benchmark Datasets: Zero-Shot Generalization}\label{sec:zero-shot-generalization}
\begin{figure}[!ht]
        \centering
         \includegraphics[width=1.125\linewidth]{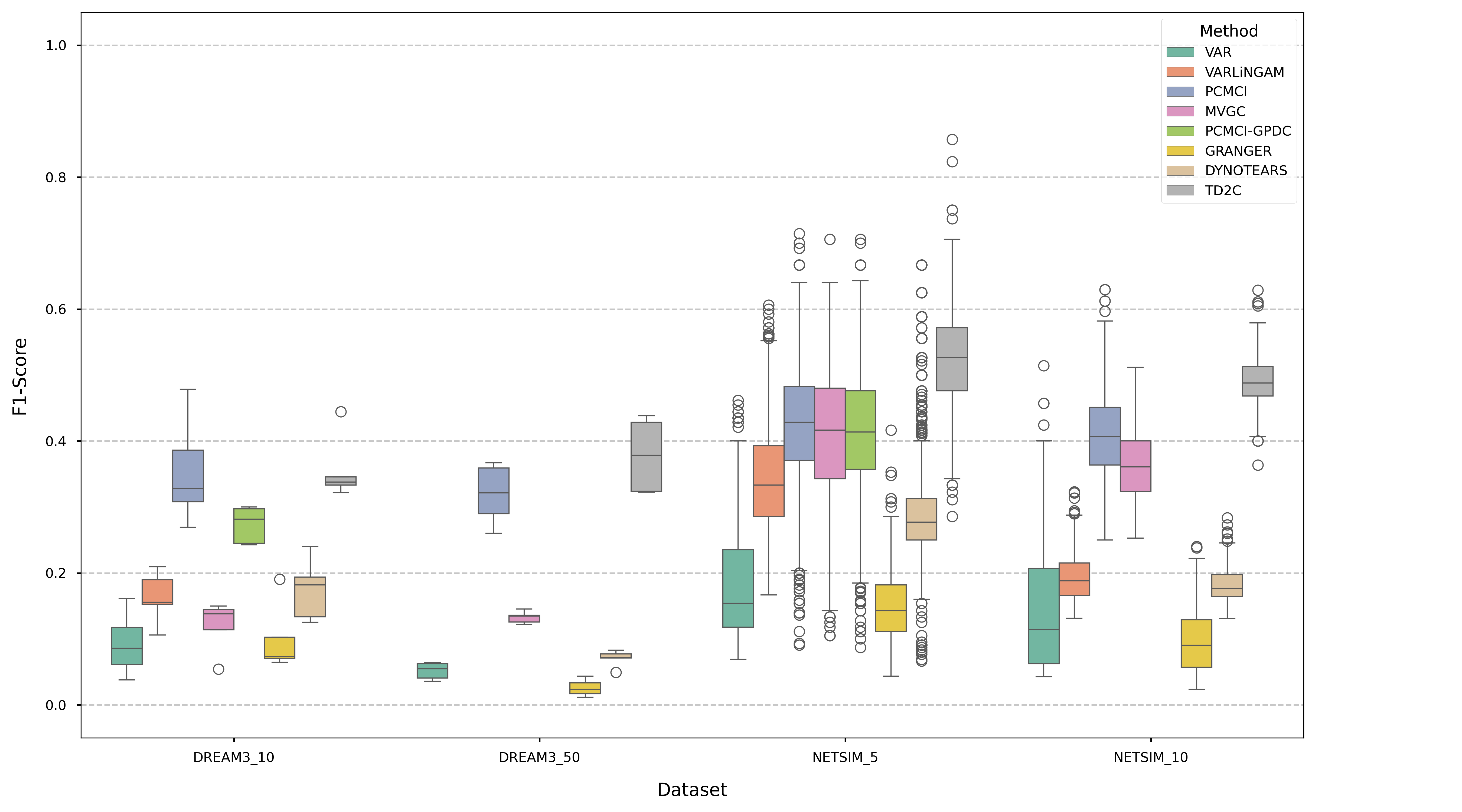}
        \caption{F1-score distribution on realistic benchmark datasets (DREAM3 and NetSim). TD2C consistently shows a high median F1-score and often a tighter distribution, indicating strong and reliable performance.}
        \label{fig:realistic-boxplot}
\end{figure}

The ultimate test for any supervised causal discovery method is its ability for zero-shot generalization: performing accurately on entirely new, real-world domains using a model trained only on synthetic data. To assess this, the TD2C classifier, trained exclusively on our aggregated synthetic dataset, was applied directly to the NetSim and DREAM3 benchmarks without any fine-tuning. The F1-Score distributions for all methods across these datasets are visualized in Figure~\ref{fig:realistic-boxplot}, providing a high-level comparison of performance. For a comprehensive analysis, full performance metrics for each benchmark are detailed in Table~\ref{tab:realistic-benchmark-details} in \ref{app:extra-results-realistic}.

As shown in the figure, TD2C consistently exhibits superior performance on the NetSim datasets, achieving high median F1-scores and often a more compact distribution, indicating both strong and reliable performance. The DREAM3 benchmarks present a more significant challenge for all methods, reflected by the overall lower F1-scores across the board. Despite this, TD2C's performance remains highly competitive, often leading the pack, particularly on the most complex DREAM3-50 dataset. It is also important to note the scalability limitations of some competing methods; the absence of PCMCI-GPDC from the NetSim-10 results highlights its challenges with long time series, while the high dimensionality of DREAM3-50 proved computationally prohibitive for both VARLiNGAM and PCMCI-GPDC.

A deeper analysis of the performance metrics, detailed in Table~\ref{tab:realistic-benchmark-details}, reveals further insights into TD2C's strengths. While the F1-score provides a combined measure, the individual Precision and Recall metrics underscore the method's balanced approach. On both NetSim and DREAM3 datasets, TD2C consistently achieves high Precision without unduly sacrificing Recall. This contrasts sharply with methods like VARLiNGAM, which often exhibit very high Recall but at the cost of extremely low Precision, rendering them impractical for discovery tasks by flooding the results with false positives. This focus on precision is even more critical on the high-dimensional DREAM3-50 benchmark, where TD2C's standout Precision is what drives its leading F1-score. Furthermore, TD2C consistently records high Balanced Accuracy scores across all datasets, indicating its effectiveness at correctly classifying both the presence and absence of causal links, thereby avoiding the bias towards the majority (no-link) class seen in some competitors.

\subsection{Statistical Validation of Results}\label{sec:results-statistical-test}

To rigorously validate the performance differences observed on the synthetic dataset, we applied the statistical testing framework described in Section~\ref{sec:statistical-tests-methodology}. Given that the F1-Score is a harmonic mean of Precision and Recall, we analyzed these two constituent metrics separately to gain a deeper understanding of each method's behavior.

\begin{figure}[!ht]
    \centering
    \begin{minipage}[b]{0.49\linewidth}
        \centering
        \includegraphics[width=\linewidth]{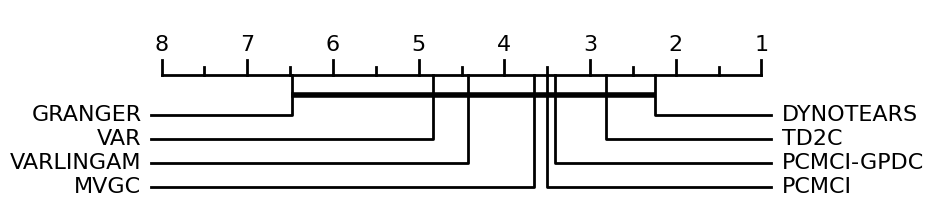}
        \centerline{(a) Precision}
        \label{fig:cd-precision}
    \end{minipage}
    \hfill     \begin{minipage}[b]{0.49\linewidth}
        \centering
        \includegraphics[width=\linewidth]{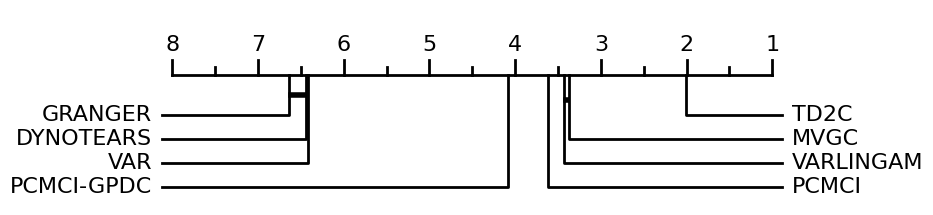}
        \centerline{(b) Recall}
        \label{fig:cd-recall}
    \end{minipage}
    \caption{Critical Difference diagrams (right is better) for (a) Precision and (b) Recall on the synthetic test set. For Precision, all methods are statistically indistinguishable. For Recall, TD2C is ranked significantly better than all other methods, demonstrating its superior ability to detect true causal links.}
    \label{fig:cd-diagrams-pr}
\end{figure}

The analysis reveals a compelling and nuanced picture. The Critical Difference (CD) diagram for Recall (Figure~\ref{fig:cd-diagrams-pr}b) shows a clear and statistically significant advantage for TD2C. It achieves the best (lowest) average rank and is not connected to any other method, indicating that its superior ability to identify true causal links is statistically robust. In sharp contrast, the CD diagram for Precision (Figure~\ref{fig:cd-diagrams-pr}a) connects all methods with a single bar, signifying that no method holds a statistically significant advantage over any other in avoiding false positives.

These findings are critical for interpreting the overall performance. TD2C's state-of-the-art F1-Score is primarily driven by its outstanding and statistically superior Recall. For completeness, the CD diagrams for all other key metrics are provided in ~\ref{app:statistical-tests}.

\subsection{Analysis of Computational Performance}\label{sec:analysis-computational-performance}

A critical aspect of any causal discovery method is its scalability. The computational complexity of TD2C is primarily driven by its pairwise analysis of all potential cause-effect-lag combinations. The total complexity can be expressed as $O(N^2 \cdot L \cdot M \log M)$, where $N$ is the number of variables, $L$ is the maximum lag considered, and $M$ is the number of time points. This complexity arises from two nested components. First, the number of candidate pairs ($O(N^2 \cdot L)$), starting from $O(N^2)$. For each pair, it considers lags up to $L$, adding a factor of $L$. Second, the dominant computational cost for analyzing a single candidate link comes from the non-parametric KSG mutual information estimator. Its complexity is governed by the underlying k-nearest neighbor search, which, for $M$ data points, scales as $O(M \log M)$ with efficient data structures like k-d trees.

While this scaling appears challenging, a key architectural advantage of TD2C is that the computation for each of the $N^2 \cdot L$ pairs is entirely independent. This makes the algorithm "embarrassingly parallel." We benchmarked the runtime of a parallelized implementation of TD2C against other methods for increasing number of variables $N$, with results shown in Table \ref{tab:runtime_summary}. These results demonstrate that parallelized TD2C is highly competitive, becoming faster than several competing methods at higher dimensions ($N > 15$).

Nevertheless, we acknowledge that the $O(N^2)$ scaling with respect to the number of variables remains a bottleneck for applications in extremely high-dimensional systems. Future work will explore strategies to mitigate this, such as employing pre-filtering steps to prune the search space or reformulating the problem to predict the entire adjacency matrix at once.

\begin{table}[!ht]
\centering
\caption{Runtime (seconds) per time series as a function of the number of variables ($N$). TD2C is shown with 1 and 50 parallel jobs to demonstrate the effectiveness of parallelization.}
\label{tab:runtime_summary}
\resizebox{\textwidth}{!}{\begin{tabular}{lrrrrrr}
\toprule
Method & $N=3$ & $N=5$ & $N=10$ & $N=15$ & $N=20$ & $N=25$ \\
\midrule
VAR & 0.023 & 0.020 & 0.032 & 0.056 & 0.157 & 0.141 \\
VARLiNGAM & 0.395 & 0.975 & 3.666 & 10.803 & 51.521 & 342.997 \\
Granger & 0.057 & 0.250 & 0.512 & 1.184 & 2.292 & 2.982 \\
DYNOTEARS & 0.013 & 0.029 & 0.016 & 0.026 & 0.023 & 0.025 \\
MultivariateGranger & 0.022 & 0.073 & 0.146 & 0.488 & 1.817 & 6.127 \\
PCMCI (ParCorr) & 0.082 & 0.309 & 0.645 & 1.399 & 2.699 & 3.777 \\
PCMCI (GPDC) & 5.686 & 9.643 & 36.563 & 66.607 & 135.967 & 191.062 \\
TD2C (1 job) & 24.041 & 65.314 & 263.030 & 612.134 & 1088.640 & 1617.625 \\
TD2C (50 jobs) & 2.590 & 3.656 & 10.558 & 39.282 & 38.932 & 57.169 \\
\bottomrule
\end{tabular}
}
\end{table}

\subsubsection{Feature Importance Analysis}\label{sec:features}
To understand the internal logic of our final model, we performed a feature importance analysis on the BalancedRandomForestClassifier after it was trained on the full synthetic training dataset. This analysis reveals which of our 63 engineered features the model found most discriminative for identifying causal links. All features and their families are detailed in Tables~\ref{tab:descriptor_formulas_math_p1} and \ref{tab:descriptor_formulas_math_p1}. For clarity and direct reference, the top 15 features are presented in Table~\ref{tab:feature-importance-synthetic} using their Python implementation names; a full mapping from the Python names to their descriptive titles and formulas is available in the \ref{app:python-mapping} Table~\ref{tab:python-mapping}.

The top two ranks are occupied by the "Correlation between the effect's prediction error and the cause" and the "Partial correlation between cause and effect residuals". This indicates that the model first learns to rely on fundamental, locally linear, relationships which are prevalent in the synthetic data. Crucially, the model does not stop there. It also assigns high importance to Information-Theoretic measures to capture more complex dynamics. Aggregate Mutual Information (MI) features such as the "Mean MI from effect's MB" and "Mean MI from cause's MB" are highly ranked. Furthermore, our novel dynamic descriptors, designed to capture non-linear and asymmetric relationships, proved to be highly valuable. The "Higher Order Cross-moment (HOC) 1,3" and our "Generalized information flow asymmetry (TE)" rank 8th and 9th, respectively, validating their inclusion and importance. \\

This analysis provides a clear rationale for TD2C's strong generalization capabilities. By learning from a varied training set to rely on a diverse feature vocabulary—spanning linear, error-based, information-theoretic, and dynamic measures—the model is not over-specialized to any single type of causal dynamic. This versatile internal logic, learned once from the synthetic data, equips the model to effectively handle the diverse and unseen challenges presented by the realistic benchmark datasets, forming the foundation of its state-of-the-art zero-shot performance.

\begin{table}[!ht]
\footnotesize
\centering
\caption{Top 15 most important features from the synthetic training dataset, listed by their Python implementation names. For a complete mapping of these names to their full mathematical formulas and descriptive titles, please refer to Table~\ref{tab:python-mapping} in \ref{app:python-mapping}}
\label{tab:feature-importance-synthetic}
\begin{tabular}{rlc}
\toprule
\textbf{Rank} & \textbf{Python Feature Name} & \textbf{Equation} \\
\midrule
1 & \texttt{errors\_correlation\_with\_inputs} & \eqref{eq:errors_correlation_with_inputs} \\
2 & \texttt{parcorr\_errors} & \eqref{eq:parcorr_errors} \\
3 & \texttt{m\_eff\_mean} & \eqref{eq:m_eff_mean} \\
4 & \texttt{coeff\_eff} & \eqref{eq:coeff_eff} \\
5 & \texttt{m\_cau\_mean} & \eqref{eq:m_cau_mean} \\
6 & \texttt{m\_cau\_interaction} & \eqref{eq:m_cau_interaction} \\
7 & \texttt{coeff\_cause} & \eqref{eq:coeff_cause} \\
8 & \texttt{HOC\_1\_3} & \eqref{eq:HOC_1_3} \\
9 & \texttt{te\_asymmetry\_diff\_1\_15} & \eqref{eq:te_asymmetry_diff_1_15} \\
10 & \texttt{eff\_cau} & \eqref{eq:eff_cau} \\
11 & \texttt{HOC\_3\_1} & \eqref{eq:HOC_3_1} \\
12 & \texttt{m\_eff\_std} & \eqref{eq:m_eff_std} \\
13 & \texttt{cau\_eff} & \eqref{eq:cau_eff} \\
14 & \texttt{mca\_mef\_eff\_std} & \eqref{eq:mca_mef_eff_std} \\
15 & \texttt{eff\_m\_cau\_mean} & \eqref{eq:eff_m_cau_mean} \\
\bottomrule
\end{tabular}
\end{table}

\section{Conclusion and Future Work}
\label{sec:conclusion_and_future_work}

In this paper, we addressed the enduring challenge of causal discovery in complex, multivariate time series. We departed from traditional approaches that rely on a single methodological family. Instead, we proposed TD2C, a supervised learning framework that reframes causal discovery as a pattern recognition problem. By engineering a comprehensive set of descriptors spanning information-theoretic, error-based, linear, and statistical domains, TD2C acts as a powerful meta-learner, capable of synthesizing evidence from the principles of many disparate causal discovery families. This "best of all worlds" approach allows it to learn the subtle, quantitative, and persistent signatures of asymmetry that causality imprints on the flow of information through a system. Our extensive experimental evaluation provided strong evidence for the efficacy of this approach. On a diverse suite of synthetic datasets, TD2C demonstrated superior generalization to unseen causal dynamics, achieving state-of-the-art performance. A rigorous statistical analysis confirmed that this was driven by a statistically significant advantage in Recall, highlighting its power in discovering true causal links. Crucially, the model trained exclusively on synthetic data exhibited remarkable zero-shot generalization to realistic benchmarks. Our feature importance analysis revealed that TD2C's robustness stems from its ability to synthesize evidence from a heterogeneous feature set, rather than relying on a single type of causal signal.

\subsection{Limitations}\label{sec:limitation}

Our TD2C method relies on several key assumptions that are common in causal discovery but may be violated in real-world applications. The method assumes causal sufficiency, the Markov condition, faithfulness, and first-order Markov self-causality. When causal sufficiency is violated due to latent confounders, TD2C may infer spurious direct causal links. Violations of the Markov condition, which can arise from cyclic dependencies or incorrect time granularity, can also cause our descriptors to fail. When faithfulness is violated, genuine causal relationships may be masked by precise parameter cancellations, leading TD2C to miss true causal links. The critical assumption of first-order Markov self-causality, while providing a significant computational advantage, may not hold in systems with higher-order dynamics, potentially leading to a misspecified Markov Blanket. Finally, the assumption of causal stationarity, which we address as a key area for future work, restricts the current model's applicability to systems where causal mechanisms are constant over time.

\subsection{Future Work}

The promising results and current limitations of TD2C define a clear and exciting path for future research.

\begin{itemize}
    \item Addressing Scalability: The most pressing challenge is the $O(N^2)$ complexity of our pairwise approach. To make TD2C applicable to truly large-scale systems, future work will focus on moving from a pairwise prediction model to one that infers the entire causal adjacency matrix in a single step, potentially leveraging graph neural network architectures.

    \item Robustness to Assumption Violations: A major research thrust will be to enhance TD2C's robustness. This includes developing descriptors sensitive to the signatures of hidden confounding and moving towards a hybrid Markov Blanket definition. Such a model could start with the simple temporal neighbors and then intelligently expand the conditioning set by selectively including a few, highly relevant cross-sectional variables, balancing computational tractability with causal accuracy.

    \item Integrating Prior Knowledge for Adaptive Learning: A particularly exciting avenue involves integrating prior domain knowledge. The supervised nature of TD2C allows known causal links to be treated as high-quality labels for paradigms like transfer learning or active learning, where a pre-trained model could be fine-tuned on domain-specific examples or query an expert for the most informative labels.

    \item Investigating Data Efficiency: A systematic study of data efficiency is needed, investigating the trade-offs between the diversity of training processes, the number of variables ($N$), and the length of the time series ($M$) on the final model's zero-shot performance. This would provide valuable guidelines for applying TD2C in new domains.

    \item Extensive Benchmarking and Real-World Application: A key future direction is to perform an extensive, large-scale validation of TD2C on standardized benchmark platforms like CauseMe \citep{munoz2020causeme}, which will test its performance across hundreds of diverse and challenging datasets. Beyond benchmarks, we aim to apply the framework to unsolved problems in fields like neuroscience or climate science, validating discovered links through domain expert knowledge.

    \item Refining the Learning Pipeline: Future efforts should involve systematically optimizing the classifier (e.g., exploring gradient boosting trees or neural networks) and key feature-engineering parameters, such as the number of neighbors ($k$) in the mutual information estimator.
\end{itemize}

A particularly critical frontier is adapting TD2C for non-stationary systems where causal mechanisms evolve over time. While the current framework assumes stationarity, its feature-based, modular design is well-suited for extension. Future work will explore methods to capture time-varying dynamics, such as implementing rolling-window analyses to track causal changes or developing fully online learning versions of the framework. An even more ambitious goal is to engineer specific descriptors sensitive to changepoints, allowing the model not just to adapt to non-stationarity but to explicitly detect and characterize the moments when the causal graph itself is restructured.

In conclusion, TD2C champions a new direction for the field. By treating causal discovery as a supervised learning problem and building a framework to integrate signals from diverse methodological families, we have shown that learning the fundamental signatures of causation is a powerful, robust, and generalizable strategy. The future of causal inference may lie not only in designing better individual tests but in creating more sophisticated meta-learners that can synthesize evidence from all available sources.

\section*{Code and Data Reproducibility}
\label{sec:reproducibility}

To ensure the transparency and reproducibility of our research, and to encourage the community to use and build upon our work to further advance the field of temporal causal discovery, the complete source code for the TD2C framework, along with the scripts used for data generation and all experiments presented in this paper, are publicly available at   \href{https://github.com/gmpal/TD2C-PP}{\texttt{https://github.com/gmpal/TD2C-PP}}.

\section*{Acknowledgements}
Gian Marco Paldino and Gianluca Bontempi are supported by the Service Public de Wallonie Recherche under grant nr 2010235–ARIAC by DigitalWallonia4.ai. Computational resources have been provided by the Consortium des Équipements de Calcul Intensif (CÉCI), funded by the Fonds de la Recherche Scientifique de Belgique (F.R.S.-FNRS) under Grant No. 2.5020.11 and by the Walloon Region.

\bibliographystyle{elsarticle-harv}
\bibliography{bibliography}

\clearpage
\appendix
\section{Complete List of TD2C Descriptors}
\label{app:full_descriptors}

\begin{table}[!ht]
\scriptsize
\centering
\caption{Mathematical Formulas for Individual Descriptors - Part 1 of 2}
\label{tab:descriptor_formulas_math_p1}
\begin{align}
& \operatorname{Corr}(\mathbf{z}_i - \operatorname{proj}_{\mathbf{MB}_i}(\mathbf{z}_i), \mathbf{z}_j - \operatorname{proj}_{\mathbf{MB}_j}(\mathbf{z}_j)) \label{eq:parcorr_errors} \\ 
& \operatorname{Corr}(\mathbf{z}_j - \operatorname{proj}_{\mathbf{MB}_i}(\mathbf{z}_j), \mathbf{z}_i) \label{eq:errors_correlation_with_inputs} \\ 
& b \text{ from } \mathbf{z}_j = b \cdot \mathbf{z}_i + \sum_{\mathbf{m}_j \in \mathbf{MB}_j} b_m \cdot \mathbf{m}_j + \epsilon \label{eq:coeff_cause} \\ 
& b \text{ from } \mathbf{z}_i = b \cdot \mathbf{z}_j + \sum_{\mathbf{m}_i \in \mathbf{MB}_i} b_m \cdot \mathbf{m}_i + \epsilon \label{eq:coeff_eff} \\ 
& \mathbb{E}[(\frac{\mathbf{z}_i-\mu_i}{\sigma_i})^3 (\frac{\mathbf{z}_j-\mu_j}{\sigma_j})^1] \label{eq:HOC_3_1} \\ 
& \mathbb{E}[(\frac{\mathbf{z}_i-\mu_i}{\sigma_i})^1 (\frac{\mathbf{z}_j-\mu_j}{\sigma_j})^2] \label{eq:HOC_1_2} \\ 
& \mathbb{E}[(\frac{\mathbf{z}_i-\mu_i}{\sigma_i})^2 (\frac{\mathbf{z}_j-\mu_j}{\sigma_j})^1] \label{eq:HOC_2_1} \\ 
& \mathbb{E}[(\frac{\mathbf{z}_i-\mu_i}{\sigma_i})^1 (\frac{\mathbf{z}_j-\mu_j}{\sigma_j})^3] \label{eq:HOC_1_3} \\ 
& \operatorname{Kurtosis}(\mathbf{z}_i) \label{eq:kurtosis_ca} \\ 
& \operatorname{Kurtosis}(\mathbf{z}_j) \label{eq:kurtosis_ef} \\ 
& \operatorname{Skewness}(\mathbf{z}_i) \label{eq:skewness_ca} \\ 
& \operatorname{Skewness}(\mathbf{z}_j) \label{eq:skewness_ef} \\ 
& \mathbb{E}_{k=1..15}\left[I(\mathbf{z}_i^{(t-1)}; \mathbf{z}_j^{(t)} | \mathbf{z}_j^{(t-k)}) - I(\mathbf{z}_j^{(t-1)}; \mathbf{z}_i^{(t)} | \mathbf{z}_i^{(t-k)})\right] \label{eq:te_asymmetry_diff_1_15} \\ 
& I(\mathbf{z}_i^{(t-1)}; \mathbf{z}_j^{(t)} | \mathbf{z}_j^{(t-1)}) \label{eq:transfer_entropy_fwd} \\ 
& I(\mathbf{z}_j^{(t-1)}; \mathbf{z}_i^{(t)} | \mathbf{z}_i^{(t-1)}) \label{eq:transfer_entropy_bwd} \\ 
& I(\mathbf{z}_i^{(t-1)}; \mathbf{z}_j^{(t)} | \mathbf{z}_j^{(t-1)}) - I(\mathbf{z}_j^{(t-1)}; \mathbf{z}_i^{(t)} | \mathbf{z}_i^{(t-1)}) \label{eq:transfer_entropy_diff} \\ 
& I(\mathbf{z}_i; \mathbf{z}_j | \mathbf{MB}_i \cap \mathbf{MB}_j) \label{eq:com_cau} \\ 
& I(\mathbf{z}_i; \mathbf{z}_j) \label{eq:cau_eff} \\ 
& I(\mathbf{z}_j; \mathbf{z}_i) \label{eq:eff_cau} \\ 
& I(\mathbf{z}_j; \mathbf{z}_i | \mathbf{MB}_j) \label{eq:eff_cau_mbeff} \\ 
& I(\mathbf{z}_i; \mathbf{z}_j | \mathbf{MB}_i) \label{eq:cau_eff_mbcau} \\ 
& I(\mathbf{z}_i^{(t-1)}; \mathbf{z}_j^{(t-1)} | \mathbf{z}_i^{(t)}) \label{eq:mca_mef_cau_parent} \\ 
& I(\mathbf{z}_i^{(t-1)}; \mathbf{z}_j^{(t+1)} | \mathbf{z}_i^{(t)}) \label{eq:mca_mef_cau_child} \\ 
& \operatorname{mean}\{I(\mathbf{m}_i; \mathbf{m}_j | \mathbf{z}_i) \mid \mathbf{m}_i \in \mathbf{MB}_i, \mathbf{m}_j \in \mathbf{MB}_j\} \label{eq:mca_mef_cau_mean} \\ 
& \operatorname{std}\{I(\mathbf{m}_i; \mathbf{m}_j | \mathbf{z}_i) \mid \mathbf{m}_i \in \mathbf{MB}_i, \mathbf{m}_j \in \mathbf{MB}_j\} \label{eq:mca_mef_cau_std} \\ 
& I(\mathbf{z}_i^{(t-1)}; \mathbf{z}_j^{(t-1)} | \mathbf{z}_j^{(t)}) \label{eq:mca_mef_eff_parent} \\ 
& I(\mathbf{z}_i^{(t-1)}; \mathbf{z}_j^{(t+1)} | \mathbf{z}_j^{(t)}) \label{eq:mca_mef_eff_child} \\ 
& \operatorname{mean}\{I(\mathbf{m}_i; \mathbf{m}_j | \mathbf{z}_j) \mid \mathbf{m}_i \in \mathbf{MB}_i, \mathbf{m}_j \in \mathbf{MB}_j\} \label{eq:mca_mef_eff_mean}
\end{align}
\end{table}

\begin{table}[!ht]
\scriptsize
\centering
\caption{(continued) Mathematical Formulas for Individual Descriptors - Part 2 of 2}
\label{tab:descriptor_formulas_math_p2}
\begin{align}
& \operatorname{std}\{I(\mathbf{m}_i; \mathbf{m}_j | \mathbf{z}_j) \mid \mathbf{m}_i \in \mathbf{MB}_i, \mathbf{m}_j \in \mathbf{MB}_j\} \label{eq:mca_mef_eff_std} \\ 
& I(\mathbf{z}_i; \mathbf{m}_j | \mathbf{z}_j), \quad \text{for } |\mathbf{MB}_j|=1 \label{eq:cau_m_eff_interaction} \\
& \operatorname{mean}\{I(\mathbf{z}_i; \mathbf{m}_j | \mathbf{z}_j) \mid \mathbf{m}_j \in \mathbf{MB}_j\} \label{eq:cau_m_eff_mean} \\ 
& \operatorname{std}\{I(\mathbf{z}_i; \mathbf{m}_j | \mathbf{z}_j) \mid \mathbf{m}_j \in \mathbf{MB}_j\} \label{eq:cau_m_eff_std} \\ 
& I(\mathbf{z}_j; \mathbf{z}_i^{(t-1)} | \mathbf{z}_i^{(t)}) \label{eq:eff_m_cau_parent} \\ 
& I(\mathbf{z}_j; \mathbf{z}_i^{(t+1)} | \mathbf{z}_i^{(t)}) \label{eq:eff_m_cau_child} \\
& \operatorname{mean}\{I(\mathbf{z}_j; \mathbf{m}_i | \mathbf{z}_i) \mid \mathbf{m}_i \in \mathbf{MB}_i\} \label{eq:eff_m_cau_mean} \\ 
& \operatorname{std}\{I(\mathbf{z}_j; \mathbf{m}_i | \mathbf{z}_i) \mid \mathbf{m}_i \in \mathbf{MB}_i\} \label{eq:eff_m_cau_std} \\  
& I(\mathbf{z}_i; \mathbf{m}_j), \quad \text{for } |\mathbf{MB}_j|=1 \label{eq:m_cau_interaction} \\ 
& \operatorname{mean}\{I(\mathbf{z}_i; \mathbf{m}_j) \mid \mathbf{m}_j \in \mathbf{MB}_j\} \label{eq:m_cau_mean} \\ 
& \operatorname{std}\{I(\mathbf{z}_i; \mathbf{m}_j) \mid \mathbf{m}_j \in \mathbf{MB}_j\} \label{eq:m_cau_std} \\ 
& I(\mathbf{z}_j; \mathbf{z}_i | \mathbf{MB}_i \cup \{\mathbf{m}_j\}), \quad \text{for } |\mathbf{MB}_j|=1 \label{eq:eff_cau_mbcau_plus_interaction} \\ 
& \operatorname{mean}\{I(\mathbf{z}_j; \mathbf{z}_i | \mathbf{MB}_i \cup \{\mathbf{m}_j\}) \mid \mathbf{m}_j \in \mathbf{MB}_j\} \label{eq:eff_cau_mbcau_plus_mean} \\ 
& \operatorname{std}\{I(\mathbf{z}_j; \mathbf{z}_i | \mathbf{MB}_i \cup \{\mathbf{m}_j\}) \mid \mathbf{m}_j \in \mathbf{MB}_j\} \label{eq:eff_cau_mbcau_plus_std} \\ 
& I(\mathbf{z}_i; \mathbf{z}_j | \mathbf{MB}_j \cup \{\mathbf{z}_i^{(t-1)}\}) \label{eq:cau_eff_mbeff_plus_parent} \\ 
& I(\mathbf{z}_i; \mathbf{z}_j | \mathbf{MB}_j \cup \{\mathbf{z}_i^{(t+1)}\}) \label{eq:cau_eff_mbeff_plus_child} \\ 
& \operatorname{mean}\{I(\mathbf{z}_i; \mathbf{z}_j | \mathbf{MB}_j \cup \{\mathbf{m}_i\}) \mid \mathbf{m}_i \in \mathbf{MB}_i\} \label{eq:cau_eff_mbeff_plus_mean} \\ 
& \operatorname{std}\{I(\mathbf{z}_i; \mathbf{z}_j | \mathbf{MB}_j \cup \{\mathbf{m}_i\}) \mid \mathbf{m}_i \in \mathbf{MB}_i\} \label{eq:cau_eff_mbeff_plus_std} \\ 
& I(\mathbf{z}_j; \mathbf{z}_i^{(t-1)}) \label{eq:m_eff_parent} \\ 
& I(\mathbf{z}_j; \mathbf{z}_i^{(t+1)}) \label{eq:m_eff_child} \\ 
& \operatorname{mean}\{I(\mathbf{z}_j; \mathbf{m}_i) \mid \mathbf{m}_i \in \mathbf{MB}_i\} \label{eq:m_eff_mean} \\ 
& \operatorname{std}\{I(\mathbf{z}_j; \mathbf{m}_i) \mid \mathbf{m}_i \in \mathbf{MB}_i\} \label{eq:m_eff_std} \\ 
& I(\mathbf{z}_i^{(t-1)}; \mathbf{z}_i^{(t+1)}) - I(\mathbf{z}_i^{(t-1)}; \mathbf{z}_i^{(t+1)}|\mathbf{z}_i^{(t)}) \label{eq:mca_mca_cau_parent} \\ 
& I(\mathbf{z}_i^{(t+1)}; \mathbf{z}_i^{(t-1)}) - I(\mathbf{z}_i^{(t+1)}; \mathbf{z}_i^{(t-1)}|\mathbf{z}_i^{(t)}) \label{eq:mca_mca_cau_child} \\ 
& \operatorname{mean}\{I(\mathbf{m}_{i,1};\mathbf{m}_{i,2}) - I(\mathbf{m}_{i,1};\mathbf{m}_{i,2}|\mathbf{z}_i) \mid \ldots\} \label{eq:mca_mca_cau_mean} \\ 
& \operatorname{std}\{I(\mathbf{m}_{i,1};\mathbf{m}_{i,2}) - I(\mathbf{m}_{i,1};\mathbf{m}_{i,2}|\mathbf{z}_i) \mid \ldots\} \label{eq:mca_mca_cau_std} \\ 
& I(\mathbf{m}_{j,1};\mathbf{m}_{j,2}) - I(\mathbf{m}_{j,1};\mathbf{m}_{j,2}|\mathbf{z}_j), \quad \text{for } |\mathbf{MB}_j|=2 \label{eq:mbe_mbe_eff_interaction} \\ 
& \operatorname{mean}\{I(\mathbf{m}_{j,1};\mathbf{m}_{j,2}) - I(\mathbf{m}_{j,1};\mathbf{m}_{j,2}|\mathbf{z}_j) \mid \ldots\} \label{eq:mbe_mbe_eff_mean} \\ 
& \operatorname{std}\{I(\mathbf{m}_{j,1};\mathbf{m}_{j,2}) - I(\mathbf{m}_{j,1};\mathbf{m}_{j,2}|\mathbf{z}_j) \mid \ldots\} \label{eq:mbe_mbe_eff_std} \\ 
& I(\mathbf{m}_i; \mathbf{m}_j | \mathbf{z}_i), \quad \text{for } |\mathbf{MB}_i|=1, |\mathbf{MB}_j|=1 \label{eq:mca_mef_cau_interaction} \\ 
& I(\mathbf{m}_i; \mathbf{m}_j | \mathbf{z}_j), \quad \text{for } |\mathbf{MB}_i|=1, |\mathbf{MB}_j|=1 \label{eq:mca_mef_eff_interaction} \\ 
& I(\mathbf{z}_j; \mathbf{m}_i | \mathbf{z}_i), \quad \text{for } |\mathbf{MB}_i|=1 \label{eq:eff_m_cau_interaction} \\ 
& I(\mathbf{z}_i; \mathbf{z}_j | \mathbf{MB}_j \cup \{\mathbf{m}_i\}), \quad \text{for } |\mathbf{MB}_i|=1 \label{eq:cau_eff_mbeff_plus_interaction} \\ 
& I(\mathbf{z}_j; \mathbf{m}_i), \quad \text{for } |\mathbf{MB}_i|=1 \label{eq:m_eff_interaction} \\ 
& I(\mathbf{m}_{i,1};\mathbf{m}_{i,2}) - I(\mathbf{m}_{i,1};\mathbf{m}_{i,2}|\mathbf{z}_i), \quad \text{for } |\mathbf{MB}_i|=2 \label{eq:mca_mca_cau_interaction}
\end{align}
\end{table}

\clearpage

\section{Python mapping of descriptors}
\label{app:python-mapping}

\begin{table}[!hbtp]
\scriptsize	
\centering
\caption{Python Feature Names and Corresponding Formula References}
\label{tab:python-mapping}
\begin{minipage}[t]{0.48\textwidth}
\centering
\begin{tabular}{|l|c|}
\hline
\textbf{Python Feature Name} & \textbf{Formula Ref.} \\ \hline
\texttt{parcorr\_errors} & (\ref{eq:parcorr_errors}) \\ \hline
\texttt{errors\_correlation\_with\_inputs} & (\ref{eq:errors_correlation_with_inputs}) \\ \hline
\texttt{coeff\_cause} & (\ref{eq:coeff_cause}) \\ \hline
\texttt{coeff\_eff} & (\ref{eq:coeff_eff}) \\ \hline
\texttt{HOC\_3\_1} & (\ref{eq:HOC_3_1}) \\ \hline
\texttt{HOC\_1\_2} & (\ref{eq:HOC_1_2}) \\ \hline
\texttt{HOC\_2\_1} & (\ref{eq:HOC_2_1}) \\ \hline
\texttt{HOC\_1\_3} & (\ref{eq:HOC_1_3}) \\ \hline
\texttt{kurtosis\_ca} & (\ref{eq:kurtosis_ca}) \\ \hline
\texttt{kurtosis\_ef} & (\ref{eq:kurtosis_ef}) \\ \hline
\texttt{skewness\_ca} & (\ref{eq:skewness_ca}) \\ \hline
\texttt{skewness\_ef} & (\ref{eq:skewness_ef}) \\ \hline
\texttt{te\_asymmetry\_diff\_1\_15} & (\ref{eq:te_asymmetry_diff_1_15}) \\ \hline
\texttt{transfer\_entropy\_fwd} & (\ref{eq:transfer_entropy_fwd}) \\ \hline
\texttt{transfer\_entropy\_bwd} & (\ref{eq:transfer_entropy_bwd}) \\ \hline
\texttt{transfer\_entropy\_diff} & (\ref{eq:transfer_entropy_diff}) \\ \hline
\texttt{com\_cau} & (\ref{eq:com_cau}) \\ \hline
\texttt{cau\_eff} & (\ref{eq:cau_eff}) \\ \hline
\texttt{eff\_cau} & (\ref{eq:eff_cau}) \\ \hline
\texttt{eff\_cau\_mbeff} & (\ref{eq:eff_cau_mbeff}) \\ \hline
\texttt{cau\_eff\_mbcau} & (\ref{eq:cau_eff_mbcau}) \\ \hline
\texttt{mca\_mef\_cau\_parent} & (\ref{eq:mca_mef_cau_parent}) \\ \hline
\texttt{mca\_mef\_cau\_child} & (\ref{eq:mca_mef_cau_child}) \\ \hline
\texttt{mca\_mef\_cau\_mean} & (\ref{eq:mca_mef_cau_mean}) \\ \hline
\texttt{mca\_mef\_cau\_std} & (\ref{eq:mca_mef_cau_std}) \\ \hline
\texttt{mca\_mef\_eff\_parent} & (\ref{eq:mca_mef_eff_parent}) \\ \hline
\texttt{mca\_mef\_eff\_child} & (\ref{eq:mca_mef_eff_child}) \\ \hline
\texttt{mca\_mef\_eff\_mean} & (\ref{eq:mca_mef_eff_mean}) \\ \hline
\texttt{mca\_mef\_eff\_std} & (\ref{eq:mca_mef_eff_std}) \\ \hline
\texttt{cau\_m\_eff\_interaction} & (\ref{eq:cau_m_eff_interaction}) \\ \hline
\texttt{cau\_m\_eff\_mean} & (\ref{eq:cau_m_eff_mean}) \\ \hline
\texttt{cau\_m\_eff\_std} & (\ref{eq:cau_m_eff_std}) \\ \hline
\end{tabular}
\end{minipage}
\hfill
\begin{minipage}[t]{0.48\textwidth}
\centering
\begin{tabular}{|l|c|}
\hline
\textbf{Python Feature Name} & \textbf{Formula Ref.} \\ \hline
\texttt{eff\_m\_cau\_parent} & (\ref{eq:eff_m_cau_parent}) \\ \hline
\texttt{eff\_m\_cau\_child} & (\ref{eq:eff_m_cau_child}) \\ \hline
\texttt{eff\_m\_cau\_mean} & (\ref{eq:eff_m_cau_mean}) \\ \hline
\texttt{eff\_m\_cau\_std} & (\ref{eq:eff_m_cau_std}) \\ \hline
\texttt{m\_cau\_interaction} & (\ref{eq:m_cau_interaction}) \\ \hline
\texttt{m\_cau\_mean} & (\ref{eq:m_cau_mean}) \\ \hline
\texttt{m\_cau\_std} & (\ref{eq:m_cau_std}) \\ \hline
\texttt{eff\_cau\_mbcau\_plus\_interaction} & (\ref{eq:eff_cau_mbcau_plus_interaction}) \\ \hline
\texttt{eff\_cau\_mbcau\_plus\_mean} & (\ref{eq:eff_cau_mbcau_plus_mean}) \\ \hline
\texttt{eff\_cau\_mbcau\_plus\_std} & (\ref{eq:eff_cau_mbcau_plus_std}) \\ \hline
\texttt{cau\_eff\_mbeff\_plus\_parent} & (\ref{eq:cau_eff_mbeff_plus_parent}) \\ \hline
\texttt{cau\_eff\_mbeff\_plus\_child} & (\ref{eq:cau_eff_mbeff_plus_child}) \\ \hline
\texttt{cau\_eff\_mbeff\_plus\_mean} & (\ref{eq:cau_eff_mbeff_plus_mean}) \\ \hline
\texttt{cau\_eff\_mbeff\_plus\_std} & (\ref{eq:cau_eff_mbeff_plus_std}) \\ \hline
\texttt{m\_eff\_parent} & (\ref{eq:m_eff_parent}) \\ \hline
\texttt{m\_eff\_child} & (\ref{eq:m_eff_child}) \\ \hline
\texttt{m\_eff\_mean} & (\ref{eq:m_eff_mean}) \\ \hline
\texttt{m\_eff\_std} & (\ref{eq:m_eff_std}) \\ \hline
\texttt{mca\_mca\_cau\_parent} & (\ref{eq:mca_mca_cau_parent}) \\ \hline
\texttt{mca\_mca\_cau\_child} & (\ref{eq:mca_mca_cau_child}) \\ \hline
\texttt{mca\_mca\_cau\_mean} & (\ref{eq:mca_mca_cau_mean}) \\ \hline
\texttt{mca\_mca\_cau\_std} & (\ref{eq:mca_mca_cau_std}) \\ \hline
\texttt{mbe\_mbe\_eff\_interaction} & (\ref{eq:mbe_mbe_eff_interaction}) \\ \hline
\texttt{mbe\_mbe\_eff\_mean} & (\ref{eq:mbe_mbe_eff_mean}) \\ \hline
\texttt{mbe\_mbe\_eff\_std} & (\ref{eq:mbe_mbe_eff_std}) \\ \hline
\texttt{mca\_mef\_cau\_interaction} & (\ref{eq:mca_mef_cau_interaction}) \\ \hline
\texttt{mca\_mef\_eff\_interaction} & (\ref{eq:mca_mef_eff_interaction}) \\ \hline
\texttt{eff\_m\_cau\_interaction} & (\ref{eq:eff_m_cau_interaction}) \\ \hline
\texttt{cau\_eff\_mbeff\_plus\_interaction} & (\ref{eq:cau_eff_mbeff_plus_interaction}) \\ \hline
\texttt{m\_eff\_interaction} & (\ref{eq:m_eff_interaction}) \\ \hline
\texttt{mca\_mca\_cau\_interaction} & (\ref{eq:mca_mca_cau_interaction}) \\ \hline
 &  \\ \hline
\end{tabular}
\end{minipage}
\end{table}

\clearpage

\section{The TD2C Training and Inference Pipeline}\label{app:pipeline}

\begin{algorithm}[!ht]
\caption{The TD2C Training and Inference Pipeline}
\label{alg:pipeline}
\begin{algorithmic}[1]
\Statex \textbf{Part 1: Training}
\State \textbf{Input:} A collection of training time series datasets $\mathcal{D} = \{D_1, \dots, D_N\}$.
\State \textbf{Input:} Corresponding ground-truth DAGs $\mathcal{G} = \{G_1, \dots, G_N\}$.
\State \textbf{Input:} A classifier model architecture, `Classifier`.
\State \textbf{Output:} A trained model, `TrainedClassifier`.

\State Initialize empty lists for feature vectors $X_{train} \gets []$ and labels $Y_{train} \gets []$.

\For{each dataset $D_k$ and its DAG $G_k$ in $(\mathcal{D}, \mathcal{G})$}
    \State Let $\mathcal{P}$ be the set of all potential time-lagged pairs $(z_i, z_j)$ in $D_k$.
    \For{each pair $(z_i, z_j) \in \mathcal{P}$}
        \State Define $\mathbf{MB}_i \gets \{z_i^{(t-1)}, z_i^{(t+1)}\}$ and $\mathbf{MB}_j \gets \{z_j^{(t-1)}, z_j^{(t+1)}\}$.
        \State Compute the full set of descriptors $\mathbf{v}_{ij}$ for the pair $(z_i, z_j)$ (Tables \ref{tab:descriptor_formulas_math_p1}, \ref{tab:descriptor_formulas_math_p2}).
        \State Summarize populations in $\mathbf{v}_{ij}$ with quantiles to create a fixed-length feature vector $\mathbf{x}_{ij}$.
        \State Set label $y_{ij} = 1$ if $z_i \to z_j$ exists in $G_k$, else $y_{ij} = 0$.
        \State Append $\mathbf{x}_{ij}$ to $X_{train}$ and $y_{ij}$ to $Y_{train}$.
    \EndFor
\EndFor
\State Train the `Classifier` on $(X_{train}, Y_{train})$ to produce `TrainedClassifier`.
\State \textbf{return} `TrainedClassifier`.

\Statex
\Statex \textbf{Part 2: Inference}
\State \textbf{Input:} An unseen time series dataset $D_{new}$.
\State \textbf{Input:} The trained model `TrainedClassifier`.
\State \textbf{Input:} A pair of interest $(z_a, z_b)$ from $D_{new}$.
\State \textbf{Output:} Causal probability $P(z_a \to z_b)$.

\State Define $\mathbf{MB}_a$ and $\mathbf{MB}_b$ from $D_{new}$ as in training.
\State Compute descriptors $\mathbf{v}_{ab}$ for the pair $(z_a, z_b)$.
\State Aggregate to create the feature vector $\mathbf{x}_{ab}$.
\State Predict probability: $p_{ab} \gets \texttt{TrainedClassifier.predict\_proba}(\mathbf{x}_{ab})$.
\State \textbf{return} $p_{ab}$.
\end{algorithmic}
\end{algorithm}

\clearpage

\section{Complete list of competitors }\label{app:competitors}

\begin{table}[!ht]
\scriptsize
\centering
\caption{Summary of benchmark causal discovery methods used for comparison.}
\label{tab:competitors}
\begin{tabular}{p{3cm} p{3cm} p{5cm}}
\toprule
\textbf{Method} & \textbf{Reference} & \textbf{Description} \\
\midrule
\multicolumn{3}{l}{\textbf{Our Method}} \\
\midrule
TD2C & This work & A two-stage method combining time series feature engineering with a supervised learning classifier to identify causal links. \\
\midrule
\multicolumn{3}{l}{\textbf{Baseline}} \\
\midrule
VAR & N/A & A naive baseline that infers causality from the statistical significance ($p-value < 0.05$) and magnitude ($|coef| > 0.1$) of coefficients in a VAR model. \\
\midrule
\multicolumn{3}{l}{\textbf{Time-Series Causality}} \\
\midrule
GRANGER & \citep{seabold2010statsmodels} & Tests if past values of one time series are useful in forecasting another. \\
MVGC & N/A & Extends Granger causality to a multivariate context, using F-tests on VAR model coefficients to test for causality conditional on all other variables.\\
\midrule
\multicolumn{3}{l}{\textbf{Constraint-Based}} \\
\midrule
PCMCI & \citep{runge2019detecting} & Identifies causal links by performing a sequence of conditional independence tests to remove spurious connections. \\
\midrule
\multicolumn{3}{l}{\textbf{Noise-Based}} \\
\midrule
VarLiNGAM & \citep{hyvarinen2010estimation} & Infers causal structure by leveraging the non-Gaussianity of error terms in a linear structural equation model. \\
\midrule
\multicolumn{3}{l}{\textbf{Score-Based}} \\
\midrule
DYNOTEARS & \citep{pamfil2020dynotears} & A score-based method that uses continuous optimization to learn a causal graph by minimizing a score function that enforces acyclicity. \\
\bottomrule
\end{tabular}
\end{table}

\clearpage

\section{Synthetic data generation details}\label{app:synthetic-data}

\begin{table}[!ht]
\resizebox{0.9\linewidth}{!}{
\begin{minipage}{1.6\linewidth}
\begin{align}
& Y_{t+1}\left[j\right]=-0.4 \frac{\left(3-\bar{Y}_t\left[\mathcal{N}_j\right]^2\right)}{\left(1+\bar{Y}_t\left[\mathcal{N}_j\right]^2\right)}+0.6 \frac{3-\left(\bar{Y}_{t-1}\left[\mathcal{N}_j\right]-0.5\right)^3}{1+\left(\bar{Y}_{t-1}\left[\mathcal{N}_j\right]-0.5\right)^4}+W_{t+1}\left[j\right] \label{eq:p1}  \\ 
& Y_{t+1}\left[j\right]=\left(0.4-2 \cos \left(40 \bar{Y}_{t-2}\left[\mathcal{N}_j\right]\right) \exp \left(-30 \bar{Y}_{t-2}\left[\mathcal{N}_j\right]^2\right)\right) \bar{Y}_{t-2}\left[\mathcal{N}_j\right]+\left(0.5-0.5 \exp \left(-50 \bar{Y}_{t-1}\left[\mathcal{N}_j\right]^2\right)\right) \bar{Y}_{t-1}\left[\mathcal{N}_j\right]+W_{t+1}\left[j\right] \label{eq:p2} \\ 
& Y_{t+1}\left[j\right]=1.5 \sin \left(\pi / 2 \bar{Y}_{t-1}\left[\mathcal{N}_j\right]\right)-\sin \left(\pi / 2 \bar{Y}_{t-2}\left[\mathcal{N}_j\right]\right)+W_{t+1}\left[j\right] \label{eq:p3}\\ 
& Y_{t+1}\left[j\right]=2 \exp \left(-0.1 \bar{Y}_t\left[\mathcal{N}_j\right]^2\right) \bar{Y}_t\left[\mathcal{N}_j\right]-\exp \left(-0.1 \bar{Y}_{t-1}\left[\mathcal{N}_j\right]^2\right) \bar{Y}_{t-1}\left[\mathcal{N}_j\right]+W_{t+1}\left[j\right] \label{eq:p4}\\ 
& Y_{t+1}\left[j\right]=-2 \bar{Y}_t\left[\mathcal{N}_j\right] I\left(\bar{Y}_t\left[\mathcal{N}_j\right]<0\right)+0.4 \bar{Y}_t\left[\mathcal{N}_j\right] I\left(\bar{Y}_t\left[\mathcal{N}_j\right]<0\right)+W_{t+1}\left[j\right]\label{eq:p5} \\ 
& Y_{t+1}\left[j\right]=0.8 \log \left(1+3 \bar{Y}_t\left[\mathcal{N}_j\right]^2\right)-0.6 \log \left(1+3 \bar{Y}_{t-2}\left[\mathcal{N}_j\right]^2\right)+W_{t+1}\left[j\right]\label{eq:p6} \\ 
& Y_{t+1}\left[j\right]=\left(0.4-2 \cos \left(40 \bar{Y}_{t-2}\left[\mathcal{N}_j\right]\right) \exp \left(-30 \bar{Y}_{t-2}\left[\mathcal{N}_j\right]^2\right)\right) \bar{Y}_{t-2}\left[\mathcal{N}_j\right]+\left(0.5-0.5 \exp \left(-50 \bar{Y}_{t-1}\left[\mathcal{N}_j\right]^2\right)\right) \bar{Y}_{t-1}\left[\mathcal{N}_j\right]+W_{t+1}\left[j\right] \label{eq:p7}\\ 
& Y_{t+1}\left[j\right]=\left(0.5-1.1 \exp \left(-50 \bar{Y}_t\left[\mathcal{N}_j\right]^2\right)\right) \bar{Y}_t\left[\mathcal{N}_j\right]+\left(0.3-0.5 \exp \left(-50 \bar{Y}_{t-2}\left[\mathcal{N}_j\right]^2\right)\right) \bar{Y}_{t-2}\left[\mathcal{N}_j\right]+W_{t+1}\left[j\right]\label{eq:p8} \\ 
& Y_{t+1}\left[j\right]=0.3 \bar{Y}_t\left[\mathcal{N}_j\right]+0.6 \bar{Y}_{t-1}\left[\mathcal{N}_j\right]+\frac{\left(0.1-0.9 \bar{Y}_t\left[\mathcal{N}_j\right]+0.8 \bar{Y}_{t-1}\left[\mathcal{N}_j\right]\right)}{\left(1+\exp \left(-10 \bar{Y}_t\left[\mathcal{N}_j\right]\right)\right)}+W_{t+1}\left[j\right]\label{eq:p9} \\ 
& Y_{t+1}\left[j\right]=\operatorname{sign}\left(\bar{Y}_t\left[\mathcal{N}_j\right]\right)+W_{t+1}\left[j\right] \label{eq:p10} \\ 
& Y_{t+1}\left[j\right]=0.8 \bar{Y}_t\left[\mathcal{N}_j\right]-\frac{0.8 \bar{Y}_t\left[\mathcal{N}_j\right]}{\left(1+\exp \left(-10 \bar{Y}_t\left[\mathcal{N}_j\right]\right)\right)}+W_{t+1}\left[j\right]\label{eq:p11} \\ 
& Y_{t+1}\left[j\right]=0.3 \bar{Y}_t\left[\mathcal{N}_j\right]+0.6 \bar{Y}_{t-1}\left[\mathcal{N}_j\right]+\frac{\left(0.1-0.9 \bar{Y}_t\left[\mathcal{N}_j\right]+0.8 \bar{Y}_{t-1}\left[\mathcal{N}_j\right]\right)}{\left(1+\exp \left(-10 \bar{Y}_t\left[\mathcal{N}_j\right]\right)\right)}+W_{t+1}\left[j\right]\label{eq:p12} \\ 
& Y_{t+1}\left[j\right]=0.38 \bar{Y}_t\left[\mathcal{N}_j\right]\left(1-\bar{Y}_{t-1}\left[\mathcal{N}_j\right]\right)+W_{t+1}\left[j\right]\label{eq:p13} \\ 
& Y_{t+1}\left[j\right]=\left\{\begin{array}{l}-0.5 \bar{Y}_t\left[\mathcal{N}_j\right] \quad \text { if } \quad \bar{Y}_t\left[\mathcal{N}_j\right]<1 \\ 0.4 \bar{Y}_t\left[\mathcal{N}_j\right]\end{array}\right.\label{eq:p14} \\ 
& Y_{t+1}\left[j\right]=\left\{\begin{array}{l}0.9 \bar{Y}_t\left[\mathcal{N}_j\right]+W_{t+1}\left[j\right] \text { if } \quad\left|\bar{Y}_t\left[\mathcal{N}_j\right]\right|<1 \\ -0.3 \bar{Y}_t\left[\mathcal{N}_j\right]+W_{t+1}\left[j\right]\end{array}\right.\label{eq:p15} \\ 
& \begin{aligned} Y_{t+1}\left[j\right] & =\left\{\begin{array}{l}-0.5 \bar{Y}_t\left[\mathcal{N}_j\right]+W_{t+1}\left[j\right] \quad \text { if } \quad x_t=1 \\ 0.4 \bar{Y}_t\left[\mathcal{N}_j\right]+W_{t+1}\left[j\right]\end{array}\right. \\ x_{t+1} & =1-x_t, x_0=1\end{aligned}\label{eq:p16} \\ 
& Y_{t+1}\left[j\right]=\sqrt{0.000019+0.846 *\left(\bar{Y}_t\left[\mathcal{N}_j\right]^2+0.3 \bar{Y}_{t-1}\left[\mathcal{N}_j\right]^2+0.2 \bar{Y}_{t-2}\left[\mathcal{N}_j\right]^2+0.1 \bar{Y}_{t-3}\left[\mathcal{N}_j\right]^2\right)} W_{t+1}\left[j\right]\label{eq:p17}\\ 
& Y_{t+1}\left[j\right] = 0.9 \cdot \bar{Y}_t\left[\mathcal{N}_j\right] + W_{t+1}\left[j\right] \label{eq:p18}\\ 
& Y_{t+1}\left[j\right] = 0.4 \cdot \bar{Y}_{t-1}\left[\mathcal{N}_j\right] + 0.6 \cdot \bar{Y}_{t-2}\left[\mathcal{N}_j\right] + W_{t+1}\left[j\right]\label{eq:p19}
\end{align}
\end{minipage}
}

\caption{Cross-sectional and temporal series: $\mathcal{N}_j$ denotes the indices of the set of time series which are parents (causes) of the $j^{th}$ time series. $\bar{y}_t\left[\mathcal{N}_j\right]$ stands for the average value of all members of $\mathcal{N}_j$ at time $t$. Generative process \eqref{eq:p17} is not considered because of its instability with the adopted initial conditions.}
\label{tab:time-series-generative-processes}
\end{table}

\clearpage

\section{Example of generated Time Series}\label{app:example-generated-time-series}
\begin{table}[!ht]
\centering
\begin{tabular}{lllll}
\toprule
0         & 1         & 2         & 3         & 4         \\ \midrule
0.712569  & 0.601736  & 1.220099  & 0.689470  & 0.387217  \\
0.856346  & 1.042727  & 1.526494  & 1.115895  & 0.909992  \\
1.380457  & 1.354174  & 1.161862  & 1.480075  & 1.408838  \\
1.515645  & 1.519205  & 0.264640  & 1.348142  & 1.488786  \\
0.628601  & 0.813913  & 0.269340  & 0.473941  & 0.589428  \\
\bottomrule
\end{tabular}
\caption{Example of generated Time Series ($N=5$) visualized in Figure \ref{fig:example-dag-a}.}
\label{tab:time_series_data_a}
\end{table}

\begin{figure}[!ht]
\centering
\includegraphics[width=0.85\linewidth]{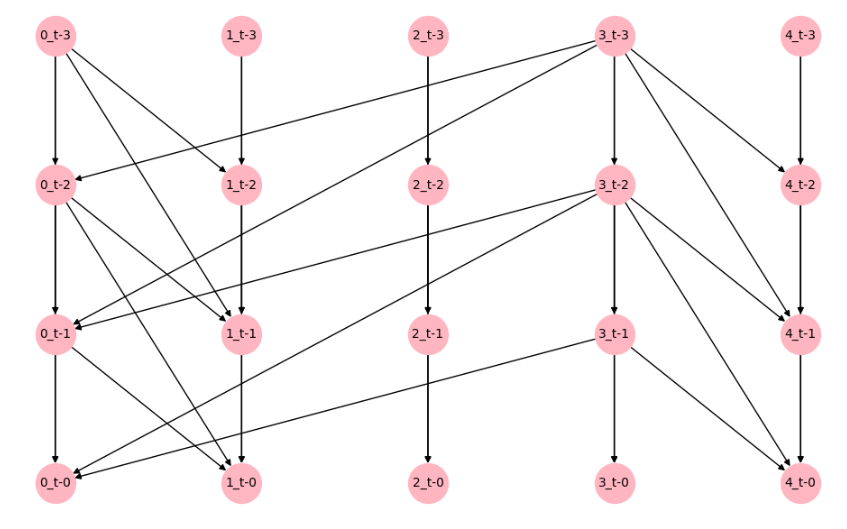}
\caption{Corresponding DAG for the example time series in Table \ref{tab:time_series_data_a}.}
\label{fig:example-dag-a}
\end{figure}

\clearpage

\section{Additional metrics on Synthetic Data}\label{app:extra-results-synthetic}

\begingroup 
\scriptsize
\setlength{\tabcolsep}{3pt} 
\begin{longtable}{llccccc}

\caption{Comprehensive comparison of eight causal discovery methods (TD2C, DYNOTEARS, Granger, MVGC, PCMCI, PCMCI-GPDC, VAR, and VARLiNGAM) evaluated across ten different synthetic generative processes (processes 2, 4, 6, 8, 10, 12, 14, 16, and 18). Performance is assessed using five key metrics: accuracy, balanced accuracy, F1-score, precision, and recall. Each cell reports the mean performance ± standard deviation across multiple experimental runs.}
\label{tab:combined_results_long} \\

\toprule
\textbf{Process} & \textbf{Method} & \textbf{Accuracy} & \textbf{Balanced Accuracy} & \textbf{F1-Score} & \textbf{Precision} & \textbf{Recall} \\
\midrule
\endfirsthead

\multicolumn{7}{c}{\tablename\ \thetable{} -- continued from previous page} \\
\toprule
\textbf{Process} & \textbf{Method} & \textbf{Accuracy} & \textbf{Balanced Accuracy} & \textbf{F1-Score} & \textbf{Precision} & \textbf{Recall} \\
\midrule
\endhead

\midrule
\multicolumn{7}{r@{}}{\textit{Continued on next page}} \\
\endfoot

\bottomrule
\endlastfoot

\multirow{8}{*}{2} & TD2C (ours) & 0.8431 \(\pm\) 0.0450 & \textbf{0.8569 \(\pm\) 0.0475} & 0.6999 \(\pm\) 0.0609 & 0.5886 \(\pm\) 0.0758 & \textbf{0.8791 \(\pm\) 0.0890} \\
 & DYNOTEARS & 0.7936 \(\pm\) 0.0282 & 0.5000 \(\pm\) 0.0000 & 0.0000 \(\pm\) 0.0000 & 0.0000 \(\pm\) 0.0000 & 0.0000 \(\pm\) 0.0000 \\
 & Granger & 0.6584 \(\pm\) 0.0777 & 0.4428 \(\pm\) 0.0575 & 0.0831 \(\pm\) 0.0919 & 0.1009 \(\pm\) 0.1283 & 0.0773 \(\pm\) 0.0848 \\
 & MVGC & 0.8740 \(\pm\) 0.0569 & 0.8322 \(\pm\) 0.0756 & \textbf{0.7143 \(\pm\) 0.1156} & 0.6980 \(\pm\) 0.1522 & 0.7611 \(\pm\) 0.1472 \\
 & PCMCI & \textbf{0.8873 \(\pm\) 0.0408} & 0.8116 \(\pm\) 0.0652 & 0.7109 \(\pm\) 0.1037 & 0.7555 \(\pm\) 0.1252 & 0.6826 \(\pm\) 0.1205 \\
 & PCMCI-GPDC & 0.8841 \(\pm\) 0.0356 & 0.7929 \(\pm\) 0.0596 & 0.6905 \(\pm\) 0.0959 & \textbf{0.7726 \(\pm\) 0.1287} & 0.6364 \(\pm\) 0.1126 \\
 & VAR & 0.8056 \(\pm\) 0.0322 & 0.5615 \(\pm\) 0.0483 & 0.2295 \(\pm\) 0.1301 & 0.6279 \(\pm\) 0.3120 & 0.1469 \(\pm\) 0.0930 \\
 & VARLiNGAM & 0.7984 \(\pm\) 0.0647 & 0.7528 \(\pm\) 0.0605 & 0.5830 \(\pm\) 0.0934 & 0.5349 \(\pm\) 0.1315 & 0.6755 \(\pm\) 0.1224 \\
\midrule

\multirow{8}{*}{4} & TD2C (ours) & 0.8813 \(\pm\) 0.0855 & 0.8257 \(\pm\) 0.0618 & 0.7299 \(\pm\) 0.0850 & 0.7641 \(\pm\) 0.1578 & 0.7366 \(\pm\) 0.1227 \\
 & DYNOTEARS & 0.8763 \(\pm\) 0.0386 & 0.7005 \(\pm\) 0.0342 & 0.5708 \(\pm\) 0.0629 & \textbf{0.9637 \(\pm\) 0.1046} & 0.4078 \(\pm\) 0.0522 \\
 & Granger & 0.5580 \(\pm\) 0.1298 & 0.4275 \(\pm\) 0.0681 & 0.1663 \(\pm\) 0.0785 & 0.1398 \(\pm\) 0.0730 & 0.2217 \(\pm\) 0.1111 \\
 & MVGC & \textbf{0.8988 \(\pm\) 0.0808} & \textbf{0.9358 \(\pm\) 0.0550} & \textbf{0.8126 \(\pm\) 0.0999} & 0.6961 \(\pm\) 0.1396 & \textbf{0.9992 \(\pm\) 0.0091} \\
 & PCMCI & 0.8921 \(\pm\) 0.0610 & 0.9159 \(\pm\) 0.0598 & 0.7894 \(\pm\) 0.0893 & 0.6782 \(\pm\) 0.1110 & 0.9563 \(\pm\) 0.0759 \\
 & PCMCI-GPDC & 0.8954 \(\pm\) 0.0461 & 0.8926 \(\pm\) 0.0533 & 0.7782 \(\pm\) 0.0717 & 0.6978 \(\pm\) 0.0874 & 0.8872 \(\pm\) 0.0814 \\
 & VAR & 0.7766 \(\pm\) 0.0499 & 0.5545 \(\pm\) 0.0628 & 0.2401 \(\pm\) 0.1449 & 0.3796 \(\pm\) 0.2306 & 0.1890 \(\pm\) 0.1304 \\
 & VARLiNGAM & 0.8210 \(\pm\) 0.1090 & 0.8860 \(\pm\) 0.0762 & 0.7089 \(\pm\) 0.1081 & 0.5604 \(\pm\) 0.1326 & 0.9974 \(\pm\) 0.0235 \\
\midrule

\multirow{8}{*}{6} & TD2C (ours) & 0.7314 \(\pm\) 0.0614 & \textbf{0.6063 \(\pm\) 0.0481} & \textbf{0.3746 \(\pm\) 0.0604} & 0.3681 \(\pm\) 0.0816 & \textbf{0.3946 \(\pm\) 0.0694} \\
 & DYNOTEARS & \textbf{0.7976 \(\pm\) 0.0316} & 0.5000 \(\pm\) 0.0000 & 0.0000 \(\pm\) 0.0000 & 0.0000 \(\pm\) 0.0000 & 0.0000 \(\pm\) 0.0000 \\
 & Granger & 0.7153 \(\pm\) 0.0587 & 0.4621 \(\pm\) 0.0397 & 0.0464 \(\pm\) 0.0625 & 0.0858 \(\pm\) 0.1464 & 0.0364 \(\pm\) 0.0508 \\
 & MVGC & 0.7728 \(\pm\) 0.0393 & 0.5344 \(\pm\) 0.0558 & 0.1668 \(\pm\) 0.1443 & 0.3064 \(\pm\) 0.2645 & 0.1317 \(\pm\) 0.1303 \\
 & PCMCI & 0.7634 \(\pm\) 0.0385 & 0.5237 \(\pm\) 0.0449 & 0.1584 \(\pm\) 0.1127 & 0.2781 \(\pm\) 0.1940 & 0.1193 \(\pm\) 0.0940 \\
 & PCMCI-GPDC & 0.7971 \(\pm\) 0.0487 & 0.5747 \(\pm\) 0.0799 & 0.2614 \(\pm\) 0.1790 & \textbf{0.4606 \(\pm\) 0.2475} & 0.1972 \(\pm\) 0.1602 \\
 & VAR & 0.7823 \(\pm\) 0.0359 & 0.4959 \(\pm\) 0.0190 & 0.0236 \(\pm\) 0.0547 & 0.0892 \(\pm\) 0.2241 & 0.0143 \(\pm\) 0.0336 \\
 & VARLiNGAM & 0.7524 \(\pm\) 0.0414 & 0.5070 \(\pm\) 0.0399 & 0.1181 \(\pm\) 0.1040 & 0.2087 \(\pm\) 0.1894 & 0.0929 \(\pm\) 0.0916 \\
 \midrule
\multirow{8}{*}{8} & TD2C (ours) & 0.7584 \(\pm\) 0.0662 & \textbf{0.6682 \(\pm\) 0.0824} & \textbf{0.4640 \(\pm\) 0.1108} & 0.4366 \(\pm\) 0.1108 & \textbf{0.5123 \(\pm\) 0.1463} \\
 & DYNOTEARS & 0.7951 \(\pm\) 0.0283 & 0.5000 \(\pm\) 0.0000 & 0.0000 \(\pm\) 0.0000 & 0.0000 \(\pm\) 0.0000 & 0.0000 \(\pm\) 0.0000 \\
 & Granger & 0.7186 \(\pm\) 0.0641 & 0.4689 \(\pm\) 0.0438 & 0.0548 \(\pm\) 0.0837 & 0.0854 \(\pm\) 0.1276 & 0.0462 \(\pm\) 0.0768 \\
 & MVGC & \textbf{0.8131 \(\pm\) 0.0598} & 0.6311 \(\pm\) 0.1094 & 0.3800 \(\pm\) 0.2191 & \textbf{0.5861 \(\pm\) 0.2600} & 0.3183 \(\pm\) 0.2245 \\
 & PCMCI & 0.8119 \(\pm\) 0.0541 & 0.6265 \(\pm\) 0.1019 & 0.3782 \(\pm\) 0.2023 & 0.5473 \(\pm\) 0.2110 & 0.3087 \(\pm\) 0.2009 \\
 & PCMCI-GPDC & 0.7979 \(\pm\) 0.0528 & 0.5851 \(\pm\) 0.0921 & 0.2860 \(\pm\) 0.1979 & 0.4721 \(\pm\) 0.2448 & 0.2212 \(\pm\) 0.1824 \\
 & VAR & 0.7909 \(\pm\) 0.0322 & 0.5102 \(\pm\) 0.0241 & 0.0588 \(\pm\) 0.0790 & 0.2811 \(\pm\) 0.3878 & 0.0341 \(\pm\) 0.0471 \\
 & VARLiNGAM & 0.7844 \(\pm\) 0.0522 & 0.6117 \(\pm\) 0.0969 & 0.3394 \(\pm\) 0.1875 & 0.4599 \(\pm\) 0.2080 & 0.3140 \(\pm\) 0.2209 \\
\midrule

\multirow{8}{*}{10} & TD2C (ours) & 0.7913 \(\pm\) 0.0634 & \textbf{0.7724 \(\pm\) 0.0653} & \textbf{0.4228 \(\pm\) 0.0666} & \textbf{0.3019 \(\pm\) 0.0651} & \textbf{0.7475 \(\pm\) 0.1305} \\
 & DYNOTEARS & 0.4160 \(\pm\) 0.1012 & 0.5380 \(\pm\) 0.0935 & 0.1960 \(\pm\) 0.0664 & 0.1156 \(\pm\) 0.0457 & 0.6903 \(\pm\) 0.1394 \\
 & Granger & 0.8473 \(\pm\) 0.0588 & 0.4849 \(\pm\) 0.0334 & 0.0341 \(\pm\) 0.0679 & 0.0492 \(\pm\) 0.1116 & 0.0322 \(\pm\) 0.0668 \\
 & MVGC & 0.8718 \(\pm\) 0.0374 & 0.5084 \(\pm\) 0.0476 & 0.0623 \(\pm\) 0.1143 & 0.0931 \(\pm\) 0.1736 & 0.0543 \(\pm\) 0.1057 \\
 & PCMCI & 0.8553 \(\pm\) 0.0306 & 0.4976 \(\pm\) 0.0471 & 0.0637 \(\pm\) 0.1089 & 0.0985 \(\pm\) 0.1902 & 0.0508 \(\pm\) 0.0869 \\
 & PCMCI-GPDC & 0.8607 \(\pm\) 0.0275 & 0.4943 \(\pm\) 0.0337 & 0.0460 \(\pm\) 0.0813 & 0.0743 \(\pm\) 0.1449 & 0.0363 \(\pm\) 0.0641 \\
 & VAR & \textbf{0.8909 \(\pm\) 0.0203} & 0.5141 \(\pm\) 0.0360 & 0.0683 \(\pm\) 0.1082 & 0.2028 \(\pm\) 0.3484 & 0.0433 \(\pm\) 0.0697 \\
 & VARLiNGAM & 0.5058 \(\pm\) 0.0827 & 0.5253 \(\pm\) 0.1046 & 0.1840 \(\pm\) 0.0793 & 0.1119 \(\pm\) 0.0527 & 0.5498 \(\pm\) 0.2010 \\
\midrule

\multirow{8}{*}{12} & TD2C (ours) & \textbf{0.9394 \(\pm\) 0.0566} & \textbf{0.9596 \(\pm\) 0.0387} & \textbf{0.8812 \(\pm\) 0.0860} & \textbf{0.8010 \(\pm\) 0.1332} & \textbf{0.9950 \(\pm\) 0.0201} \\
 & DYNOTEARS & 0.8136 \(\pm\) 0.0355 & 0.5205 \(\pm\) 0.0388 & 0.0696 \(\pm\) 0.1262 & 0.2917 \(\pm\) 0.4564 & 0.0411 \(\pm\) 0.0777 \\
 & Granger & 0.5994 \(\pm\) 0.1100 & 0.3994 \(\pm\) 0.0585 & 0.0692 \(\pm\) 0.0560 & 0.0662 \(\pm\) 0.0574 & 0.0785 \(\pm\) 0.0664 \\
 & MVGC & 0.8526 \(\pm\) 0.0620 & 0.8213 \(\pm\) 0.0610 & 0.6752 \(\pm\) 0.0854 & 0.6264 \(\pm\) 0.1382 & 0.7698 \(\pm\) 0.1261 \\
 & PCMCI & 0.8718 \(\pm\) 0.0473 & 0.8116 \(\pm\) 0.0572 & 0.6860 \(\pm\) 0.0824 & 0.6771 \(\pm\) 0.1156 & 0.7127 \(\pm\) 0.1084 \\
 & PCMCI-GPDC & 0.8614 \(\pm\) 0.0401 & 0.7533 \(\pm\) 0.0472 & 0.6187 \(\pm\) 0.0802 & 0.6808 \(\pm\) 0.1203 & 0.5760 \(\pm\) 0.0807 \\
 & VAR & 0.8016 \(\pm\) 0.0359 & 0.5406 \(\pm\) 0.0404 & 0.1772 \(\pm\) 0.1120 & 0.4630 \(\pm\) 0.2939 & 0.1148 \(\pm\) 0.0772 \\
 & VARLiNGAM & 0.7921 \(\pm\) 0.0750 & 0.7392 \(\pm\) 0.0602 & 0.5565 \(\pm\) 0.0878 & 0.5056 \(\pm\) 0.1333 & 0.6527 \(\pm\) 0.1034 \\
\midrule

\multirow{8}{*}{14} & TD2C (ours) & 0.8977 \(\pm\) 0.0321 & 0.9343 \(\pm\) 0.0297 & 0.6549 \(\pm\) 0.0623 & 0.4949 \(\pm\) 0.0684 & \textbf{0.9801 \(\pm\) 0.0496} \\
 & DYNOTEARS & 0.9034 \(\pm\) 0.0155 & 0.5000 \(\pm\) 0.0000 & 0.0000 \(\pm\) 0.0000 & 0.0000 \(\pm\) 0.0000 & 0.0000 \(\pm\) 0.0000 \\
 & Granger & 0.7849 \(\pm\) 0.0636 & 0.4560 \(\pm\) 0.0501 & 0.0410 \(\pm\) 0.0756 & 0.0358 \(\pm\) 0.0698 & 0.0498 \(\pm\) 0.0896 \\
 & MVGC & 0.9280 \(\pm\) 0.0510 & 0.9386 \(\pm\) 0.0482 & 0.7406 \(\pm\) 0.1391 & 0.6282 \(\pm\) 0.1870 & 0.9516 \(\pm\) 0.0797 \\
 & PCMCI & \textbf{0.9529 \(\pm\) 0.0241} & \textbf{0.9651 \(\pm\) 0.0314} & \textbf{0.8039 \(\pm\) 0.0882} & \textbf{0.6904 \(\pm\) 0.1212} & \textbf{0.9801 \(\pm\) 0.0547} \\
 & PCMCI-GPDC & 0.9511 \(\pm\) 0.0261 & 0.9579 \(\pm\) 0.0376 & 0.7973 \(\pm\) 0.0923 & 0.6878 \(\pm\) 0.1246 & 0.9663 \(\pm\) 0.0665 \\
 & VAR & 0.9044 \(\pm\) 0.0196 & 0.5737 \(\pm\) 0.0705 & 0.2294 \(\pm\) 0.1810 & 0.4528 \(\pm\) 0.3576 & 0.1650 \(\pm\) 0.1437 \\
 & VARLiNGAM & 0.9264 \(\pm\) 0.0427 & 0.9386 \(\pm\) 0.0448 & 0.7285 \(\pm\) 0.1156 & 0.6052 \(\pm\) 0.1524 & 0.9537 \(\pm\) 0.0809 \\
\midrule

\multirow{8}{*}{16} & TD2C (ours) & 0.8426 \(\pm\) 0.0361 & \textbf{0.7761 \(\pm\) 0.0532} & \textbf{0.4722 \(\pm\) 0.0580} & \textbf{0.3620 \(\pm\) 0.0557} & \textbf{0.6918 \(\pm\) 0.0925} \\
 & DYNOTEARS & \textbf{0.8992 \(\pm\) 0.0138} & 0.5000 \(\pm\) 0.0000 & 0.0000 \(\pm\) 0.0000 & 0.0000 \(\pm\) 0.0000 & 0.0000 \(\pm\) 0.0000 \\
 & Granger & 0.8418 \(\pm\) 0.0422 & 0.4707 \(\pm\) 0.0230 & 0.0072 \(\pm\) 0.0321 & 0.0086 \(\pm\) 0.0397 & 0.0063 \(\pm\) 0.0278 \\
 & MVGC & 0.8418 \(\pm\) 0.0365 & 0.5070 \(\pm\) 0.0501 & 0.0884 \(\pm\) 0.1050 & 0.0993 \(\pm\) 0.1203 & 0.0879 \(\pm\) 0.1131 \\
 & PCMCI & 0.8191 \(\pm\) 0.0287 & 0.4831 \(\pm\) 0.0428 & 0.0603 \(\pm\) 0.0825 & 0.0604 \(\pm\) 0.0829 & 0.0625 \(\pm\) 0.0880 \\
 & PCMCI-GPDC & 0.8278 \(\pm\) 0.0280 & 0.4810 \(\pm\) 0.0386 & 0.0491 \(\pm\) 0.0755 & 0.0549 \(\pm\) 0.0872 & 0.0466 \(\pm\) 0.0720 \\
 & VAR & 0.8824 \(\pm\) 0.0203 & 0.4945 \(\pm\) 0.0162 & 0.0130 \(\pm\) 0.0464 & 0.0338 \(\pm\) 0.1462 & 0.0087 \(\pm\) 0.0310 \\
 & VARLiNGAM & 0.8161 \(\pm\) 0.0489 & 0.4765 \(\pm\) 0.0460 & 0.0465 \(\pm\) 0.0812 & 0.0480 \(\pm\) 0.0908 & 0.0514 \(\pm\) 0.0929 \\
\midrule

\multirow{8}{*}{18} & TD2C (ours) & \textbf{0.9941 \(\pm\) 0.0113} & \textbf{0.9967 \(\pm\) 0.0064} & \textbf{0.9757 \(\pm\) 0.0453} & \textbf{0.9560 \(\pm\) 0.0803} & \textbf{1.0000 \(\pm\) 0.0000} \\
 & DYNOTEARS & 0.9079 \(\pm\) 0.0189 & 0.5338 \(\pm\) 0.0648 & 0.1026 \(\pm\) 0.1866 & 0.2535 \(\pm\) 0.4280 & 0.0686 \(\pm\) 0.1314 \\
 & Granger & 0.7156 \(\pm\) 0.1255 & 0.4261 \(\pm\) 0.0791 & 0.0447 \(\pm\) 0.0775 & 0.0345 \(\pm\) 0.0629 & 0.0691 \(\pm\) 0.1148 \\
 & MVGC & 0.9270 \(\pm\) 0.0593 & 0.9594 \(\pm\) 0.0332 & 0.7577 \(\pm\) 0.1420 & 0.6312 \(\pm\) 0.1896 & \textbf{1.0000 \(\pm\) 0.0000} \\
 & PCMCI & 0.9527 \(\pm\) 0.0259 & 0.9738 \(\pm\) 0.0144 & 0.8123 \(\pm\) 0.0860 & 0.6929 \(\pm\) 0.1250 & \textbf{1.0000 \(\pm\) 0.0000} \\
 & PCMCI-GPDC & 0.9581 \(\pm\) 0.0240 & 0.9768 \(\pm\) 0.0134 & 0.8299 \(\pm\) 0.0813 & 0.7174 \(\pm\) 0.1191 & \textbf{1.0000 \(\pm\) 0.0000} \\
 & VAR & 0.8993 \(\pm\) 0.0227 & 0.5834 \(\pm\) 0.0594 & 0.2595 \(\pm\) 0.1480 & 0.4859 \(\pm\) 0.3192 & 0.1913 \(\pm\) 0.1203 \\
 & VARLiNGAM & 0.9388 \(\pm\) 0.0502 & 0.9660 \(\pm\) 0.0279 & 0.7863 \(\pm\) 0.1368 & 0.6685 \(\pm\) 0.1857 & \textbf{1.0000 \(\pm\) 0.0000} \\
\bottomrule
\end{longtable}

\endgroup

\clearpage

\section{Additional metrics on Realistic Data}\label{app:extra-results-realistic}

\begingroup 
\scriptsize
\setlength{\tabcolsep}{3pt} 
\begin{longtable}{llccccc}
\caption{Detailed performance on realistic benchmark datasets. The model was trained on synthetic data and evaluated in a zero-shot setting. Scores are presented as mean \(\pm\) standard deviation. Best performance for each metric within a dataset is highlighted in \textbf{bold}.}
\label{tab:realistic-benchmark-details} \\

\toprule
\textbf{Dataset} & \textbf{Method} & \textbf{Accuracy} & \textbf{Balanced Accuracy} & \textbf{F1-Score} & \textbf{Precision} & \textbf{Recall} \\
\midrule
\endfirsthead

\multicolumn{7}{c}{\tablename\ \thetable{} -- continued from previous page} \\
\toprule
\textbf{Dataset} & \textbf{Method} & \textbf{Accuracy} & \textbf{Balanced Accuracy} & \textbf{F1-Score} & \textbf{Precision} & \textbf{Recall} \\
\midrule
\endhead

\bottomrule
\endlastfoot

\multirow{8}{*}{DREAM3\_10} & TD2C (ours) & 0.8267 \(\pm\) 0.0167 & \textbf{0.7066 \(\pm\) 0.0390} & \textbf{0.3566 \(\pm\) 0.0499} & 0.2728 \(\pm\) 0.0721 & 0.5582 \(\pm\) 0.1022 \\
 & DYNOTEARS & 0.8800 \(\pm\) 0.0217 & 0.5477 \(\pm\) 0.0191 & 0.1747 \(\pm\) 0.0470 & 0.2392 \(\pm\) 0.1043 & 0.1436 \(\pm\) 0.0299 \\
 & GRANGER & 0.8213 \(\pm\) 0.0571 & 0.4878 \(\pm\) 0.0433 & 0.1002 \(\pm\) 0.0603 & 0.0814 \(\pm\) 0.0709 & 0.0837 \(\pm\) 0.0690 \\
 & MVGC & 0.8560 \(\pm\) 0.0563 & 0.5105 \(\pm\) 0.0173 & 0.1201 \(\pm\) 0.0445 & 0.1030 \(\pm\) 0.0710 & 0.0928 \(\pm\) 0.0682 \\
 & PCMCI & \textbf{0.8840 \(\pm\) 0.0215} & 0.6557 \(\pm\) 0.0599 & 0.3538 \(\pm\) 0.0814 & \textbf{0.3660 \(\pm\) 0.1042} & 0.3754 \(\pm\) 0.1312 \\
 & PCMCI-GPDC & 0.8453 \(\pm\) 0.0219 & 0.6181 \(\pm\) 0.0363 & 0.2733 \(\pm\) 0.0279 & 0.2350 \(\pm\) 0.0262 & 0.3409 \(\pm\) 0.0873 \\
 & VAR & 0.8633 \(\pm\) 0.0355 & 0.5006 \(\pm\) 0.0272 & 0.0926 \(\pm\) 0.0529 & 0.0974 \(\pm\) 0.0740 & 0.0608 \(\pm\) 0.0543 \\
 & VARLiNGAM & 0.2160 \(\pm\) 0.0400 & 0.5078 \(\pm\) 0.0546 & 0.1626 \(\pm\) 0.0396 & 0.0901 \(\pm\) 0.0239 & \textbf{0.8630 \(\pm\) 0.1103} \\
\midrule

\multirow{6}{*}{DREAM3\_50} & TD2C (ours) & \textbf{0.9763 \(\pm\) 0.0055} & 0.6722 \(\pm\) 0.0516 & \textbf{0.3782 \(\pm\) 0.0553} & \textbf{0.4460 \(\pm\) 0.0856} & 0.3539 \(\pm\) 0.1055 \\
 & DYNOTEARS & 0.9607 \(\pm\) 0.0047 & 0.5276 \(\pm\) 0.0120 & 0.0705 \(\pm\) 0.0128 & 0.0718 \(\pm\) 0.0133 & 0.0749 \(\pm\) 0.0263 \\
 & GRANGER & 0.8450 \(\pm\) 0.0626 & 0.4785 \(\pm\) 0.0108 & 0.0257 \(\pm\) 0.0129 & 0.0149 \(\pm\) 0.0072 & 0.0962 \(\pm\) 0.0570 \\
 & MVGC & 0.9206 \(\pm\) 0.0130 & 0.6156 \(\pm\) 0.0302 & 0.1327 \(\pm\) 0.0093 & 0.0869 \(\pm\) 0.0103 & 0.2968 \(\pm\) 0.0620 \\
 & PCMCI & 0.9627 \(\pm\) 0.0098 & \textbf{0.6998 \(\pm\) 0.0561} & 0.3195 \(\pm\) 0.0454 & 0.2623 \(\pm\) 0.0324 & \textbf{0.4246 \(\pm\) 0.1102} \\
 & VAR & 0.9654 \(\pm\) 0.0069 & 0.5154 \(\pm\) 0.0068 & 0.0513 \(\pm\) 0.0127 & 0.0624 \(\pm\) 0.0200 & 0.0452 \(\pm\) 0.0136 \\
\midrule

\multirow{8}{*}{NETSIM\_5} & TD2C (ours) & \textbf{0.8533 \(\pm\) 0.0525} & \textbf{0.7349 \(\pm\) 0.0443} & \textbf{0.5215 \(\pm\) 0.0745} & \textbf{0.5113 \(\pm\) 0.1459} & 0.5727 \(\pm\) 0.1135 \\
 & DYNOTEARS & 0.4629 \(\pm\) 0.1851 & 0.5969 \(\pm\) 0.0767 & 0.3000 \(\pm\) 0.0933 & 0.2085 \(\pm\) 0.1390 & 0.7806 \(\pm\) 0.2003 \\
 & GRANGER & 0.7271 \(\pm\) 0.1311 & 0.4746 \(\pm\) 0.0521 & 0.1481 \(\pm\) 0.0521 & 0.1048 \(\pm\) 0.1279 & 0.1290 \(\pm\) 0.1348 \\
 & MVGC & 0.7795 \(\pm\) 0.0968 & 0.6792 \(\pm\) 0.0654 & 0.4089 \(\pm\) 0.0986 & 0.3590 \(\pm\) 0.1253 & 0.5417 \(\pm\) 0.1579 \\
 & PCMCI & 0.7732 \(\pm\) 0.0896 & 0.6953 \(\pm\) 0.0735 & 0.4240 \(\pm\) 0.0961 & 0.3450 \(\pm\) 0.1011 & 0.5886 \(\pm\) 0.1241 \\
 & PCMCI-GPDC & 0.7662 \(\pm\) 0.0847 & 0.6916 \(\pm\) 0.0742 & 0.4133 \(\pm\) 0.0953 & 0.3323 \(\pm\) 0.0996 & 0.5894 \(\pm\) 0.1417 \\
 & VAR & 0.7969 \(\pm\) 0.0488 & 0.5076 \(\pm\) 0.0509 & 0.1807 \(\pm\) 0.0792 & 0.1474 \(\pm\) 0.1457 & 0.1116 \(\pm\) 0.1153 \\
 & VARLiNGAM & 0.5504 \(\pm\) 0.1468 & 0.6680 \(\pm\) 0.0854 & 0.3448 \(\pm\) 0.0773 & 0.2227 \(\pm\) 0.0647 & \textbf{0.8288 \(\pm\) 0.1336} \\
\midrule

\multirow{7}{*}{NETSIM\_10} & TD2C (ours) & \textbf{0.9265 \(\pm\) 0.0119} & 0.7306 \(\pm\) 0.0180 & \textbf{0.4918 \(\pm\) 0.0414} & \textbf{0.4891 \(\pm\) 0.0775} & 0.5027 \(\pm\) 0.0384 \\
 & DYNOTEARS & 0.4759 \(\pm\) 0.0802 & 0.6358 \(\pm\) 0.0560 & 0.1826 \(\pm\) 0.0284 & 0.1030 \(\pm\) 0.0175 & 0.8217 \(\pm\) 0.0816 \\
 & GRANGER & 0.8446 \(\pm\) 0.0456 & 0.5008 \(\pm\) 0.0361 & 0.0966 \(\pm\) 0.0454 & 0.0749 \(\pm\) 0.0555 & 0.1011 \(\pm\) 0.0822 \\
 & MVGC & 0.8561 \(\pm\) 0.0383 & 0.7235 \(\pm\) 0.0398 & 0.3627 \(\pm\) 0.0523 & 0.2747 \(\pm\) 0.0629 & 0.5693 \(\pm\) 0.1024 \\
 & PCMCI & 0.8772 \(\pm\) 0.0285 & \textbf{0.7513 \(\pm\) 0.0414} & 0.4144 \(\pm\) 0.0706 & 0.3196 \(\pm\) 0.0716 & 0.6048 \(\pm\) 0.0767 \\
 & VAR & 0.9004 \(\pm\) 0.0191 & 0.5299 \(\pm\) 0.0466 & 0.1475 \(\pm\) 0.1039 & 0.1801 \(\pm\) 0.1716 & 0.0990 \(\pm\) 0.0849 \\
 & VARLiNGAM & 0.4674 \(\pm\) 0.1114 & 0.6654 \(\pm\) 0.0660 & 0.1962 \(\pm\) 0.0377 & 0.1107 \(\pm\) 0.0239 & \textbf{0.8956 \(\pm\) 0.0675} \\

\end{longtable}

\endgroup

\clearpage

\section{Additional Statistical Tests}\label{app:statistical-tests}
\begin{figure}[!ht]
    \centering
        \begin{minipage}[b]{0.49\linewidth}
        \centering
        \includegraphics[width=\linewidth]{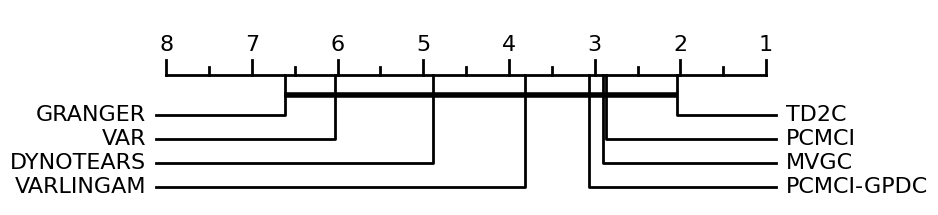}         \centerline{(a) F1-Score}
        \label{fig:cd-f1-score}
    \end{minipage}
    \hfill     \begin{minipage}[b]{0.49\linewidth}
        \centering
        \includegraphics[width=\linewidth]{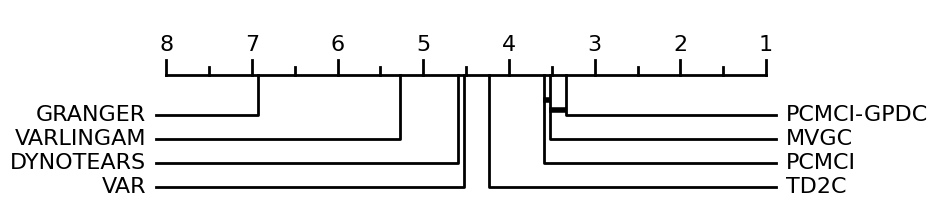}         \centerline{(b) Accuracy}
        \label{fig:cd-accuracy}
    \end{minipage}

    \vspace{1cm} 
        \begin{minipage}[b]{0.6\linewidth}         \centering
        \includegraphics[width=\linewidth]{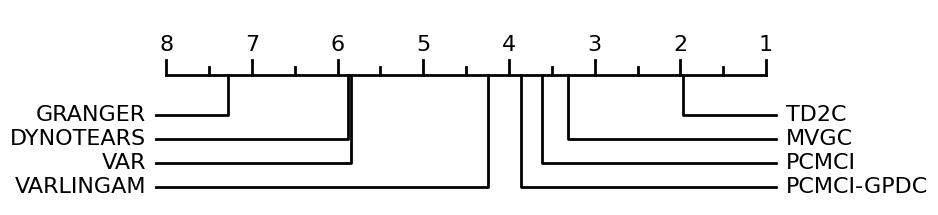}         \centerline{(c) Balanced Accuracy}
        \label{fig:cd-balanced-accuracy}
    \end{minipage}

    \caption{Critical Difference diagrams for additional performance metrics (right is better). 
    (a) For F1-Score, TD2C achieves the best average rank but is statistically grouped with other top-tier methods (PCMCI, MVGC, PCMCI-GPDC). 
    (b) For Accuracy, other methods like PCMCI-GPDC rank highest, illustrating the limitations of this metric for imbalanced discovery tasks. 
    (c) For Balanced Accuracy, TD2C again leads in rank, though its performance is statistically similar to a broad group of competitors.}
    \label{fig:cd-diagrams-appendix}
\end{figure}

Following the analysis of Precision and Recall in the main text, this appendix provides the Critical Difference (CD) diagrams for F1-Score, Accuracy, and Balanced Accuracy (Figure~\ref{fig:cd-diagrams-appendix}).

\end{document}